\pdfoutput=1
\documentclass[11pt]{article}

\usepackage{marvosym}
\usepackage[preprint]{acl}

\usepackage{comment}
\usepackage[utf8]{inputenc}
\usepackage{times}
\usepackage{amsmath,amssymb}
\usepackage{algorithm}
\usepackage{algpseudocode}

\usepackage{arydshln}
\usepackage{booktabs}
\usepackage{latexsym}
\usepackage{multirow}
\usepackage[T1]{fontenc}
\usepackage{times}
\usepackage{textcomp}
\algnewcommand\And{\textbf{and}}
\usepackage[utf8]{inputenc}
\usepackage{microtype}
\usepackage{tabulary,capt-of}
\usepackage{CJKutf8}
\newcommand{\heart}{\ensuremath\heartsuit}
\usepackage{inconsolata}
\usepackage{colortbl}
\usepackage{graphicx}
\usepackage{subfigure}
\usepackage{subcaption}
\definecolor{r7}{RGB}{132,159,219}
\definecolor{r20}{RGB}{252, 252, 255}
\definecolor{d19}{RGB}{250,233, 176}
\definecolor{e20}{RGB}{254, 254, 253}
\definecolor{y5}{RGB}{255,217,96}

\newlength\myindent
\title{How to Learn in a Noisy World? \\  \textit{Self}-Correcting the Real-World Data Noise in Machine Translation}

\author{
    Yan Meng \quad Di Wu \quad Christof Monz \\
    Language Technology Lab\\
    University of Amsterdam\\
    \texttt{\{y.meng\}@uva.nl}
}

\begin{document}
\maketitle
\begin{abstract}

The massive amounts of web-mined parallel data contain large amounts of noise.
Semantic misalignment, as the primary source of the noise, poses a challenge for training machine translation systems.
In this paper, we first introduce a process for simulating misalignment controlled by semantic similarity, which closely resembles misaligned sentences in real-world web-crawled corpora. 
Under our simulated misalignment noise settings, we quantitatively analyze its impact on machine translation and demonstrate the limited effectiveness of widely used pre-filters for noise detection. 
This underscores the necessity of more fine-grained ways to handle hard-to-detect misalignment noise.
With an observation of the increasing reliability of the model's self-knowledge for distinguishing misaligned and clean data at the token level, we propose \textit{self-correction}---an approach
that gradually increases trust in the model’s self-knowledge to correct the training supervision.
% that gradually increases trust in the model’s self-knowledge to correct the training supervision from the ground truth.
Comprehensive experiments show that our method significantly improves translation performance both in the presence of simulated misalignment noise and when applied to real-world, noisy web-mined datasets, across a range of translation tasks.

\end{abstract}

\section{Introduction}  

The success of machine translation (MT) models is mainly due to the availability of large amounts of web-crawled data.
However, publicly available web-mined parallel corpora such as CCAligned \citep{ElKishky2019AMC}, WikiMatrix \citep{Schwenk2017LearningJM} and ParaCrawl \citep{banon-etal-2020-paracrawl} are shown to be noisy \citep{kreutzer-etal-2022-quality, Ranathunga2024QualityDM}. 
The notable performance drop in NMT when training with injected synthetic noise \citep{khayrallah-koehn-2018-impact} or fine-tuning with CCAligned \citep{lee-etal-2022-pre} indicates the importance of improving the model's robustness when training with the noisy corpus.

\begin{comment}
\begin{figure}
    \centering
    \includegraphics[width=0.5\textwidth]{latex/image/intro.pdf}
    \caption{\small An illustration of our self-correction method. When the model's translation is superior than the human reference, e.g., albergo means ``hotel'' instead of ``house'', we self-correct the ground-truth data by the model's prediction and learn towards the new revised target. }
    \label{fig:intro}
   \vspace{-0.3cm}
\end{figure}
\end{comment}

Given a noisy training dataset, a common and straightforward approach to mitigate the impact of noisy data is to filter low-quality training samples \citep{herold-etal-2022-detecting, bane-etal-2022-comparison}. 
However, in practice, large amounts of misalignments still exist in \textit{pre-filtered} web-mined datasets \citep{kreutzer-etal-2022-quality}. 
This is because real-world misaligned sentences often share partial meanings, making them appear as seemingly parallel data, increasing the difficulty for pre-filters to detect them. 
To quantitatively analyze such hard-to-detect real-world misalignments, we design a process to simulate it controlled by semantic similarity. 
Unlike earlier works \citep{khayrallah-koehn-2018-impact, herold-etal-2022-detecting, Li2023ErrorNT} that generate misaligned bitext by random shuffling---an approach that is both unrealistic and easy-to-detect---our simulated misalignments closely resemble real-world noise and challenge widely-used pre-filters, such as LASER \citep{Artetxe2018MassivelyMS} and COMET \citep{rei-etal-2020-COMET}.

% Given a noisy training dataset, a common and straightforward approach to mitigate the impact of noisy data is to filter low-quality training samples \citep{herold-etal-2022-detecting, bane-etal-2022-comparison}.  
% However, in practice, large amounts of misaligned samples still exist in \textit{pre-filtered} web-mined datasets \citep{kreutzer-etal-2022-quality}. 
% Earlier works \citep{briakou-carpuat-2020-detecting, herold-etal-2022-detecting, Li2023ErrorNT} study the impact of misalignment on machine translation. 
% However, they generate synthetic misaligned bitext by random shuffling or sample perturbation, which is unrealistic.
% The real-world misaligned targets are natural sentences that contain partially shared meanings, which disguise them as seemingly parallel data and are hard to detect. 
% To quantitatively analyze such real-world misalignment, we design a process to simulate it controlled by semantic similarity. 
% Similar to the real-world setting, we show that our simulated misalignment also challenges widely-used pre-filters, such as LASER \citep{Artetxe2018MassivelyMS} and COMET \citep{rei-etal-2020-COMET}. 

Under our simulated noise settings, we evaluate a type of approach that could potentially handle misalignment noise: Data truncation \citep{kang-hashimoto-2020-improved, Li2023ErrorNT, flores-cohan-2024-benefits}, which ignores losses at the token level during training when there is a relatively large discrepancy between the model's prediction and the ground truth.  
Although promising, we observe that truncation methods are sensitive to varying levels of misalignment noise. 
For example, for low-resource corpora with a high misalignment rate, truncation methods even \textit{degrade} the translation performance (see Section~\ref{sec:real}). 
We argue that the noisy low-resource setting prevents the model from acquiring sufficient correct knowledge, resulting in an inaccurate removal of clean and useful ground-truth data.
Moreover, truncation methods start to ignore potential data noise from an early training time, which overlooks the increasing reliability of the model's prediction over time. 

% Under our simulated noise setting, we evaluate an alternative type of approach to handle data noise: Data truncation \citep{kang-hashimoto-2020-improved, Li2023ErrorNT}, which focuses on the model training dynamics and ignores losses at the token level when there is a relatively large inconsistency between the model's prediction and the ground truth during training.  
% However, we observe that truncation methods are sensitive to varying levels of simulated misalignment noise. 
% For example, in low-resource training corpus with a high misalignment rate, truncation methods \textit{degrade} the translation performance (shown in Section \ref{sec:real}). 
% We argue that the noisy low-resource setting restricts the model from acquiring sufficient correct knowledge, resulting in an inaccurate removal of clean and useful ground-truth data.
% Moreover, truncation methods start to ignore potential data noise from an early training time, which overlooks the increasing reliability of the model's prediction over time. 

To overcome these limitations, we propose an approach called \textit{self-correction}, which leverages the model's self-knowledge to correct noise during training while maintaining supervision from the ground truth to avoid discarding useful training information. 
To adapt to the model's changing reliability, we set a dynamic schedule to gradually increase the trust in its output. 
During the early stages of training, we place greater trust in the reference over the model's predictions. 
As the model acquires more knowledge, we progressively use the model's predictions to revise the ground truth. 

%training supervision gradually relies on the model's self-knowledge.
% To overcome these limitations, we propose an approach called self-correction. This approach leverages the model's predictions to self-correct noise during training while maintaining supervision from the ground truth to avoid discarding useful training information. 
% To adapt to the changing reliability of the model's self-knowledge, we set a dynamic schedule to gradually increase the trust in the model's predictions to revise the training supervision from the ground truth. 
% In the early stage of training, the model is not well trained, so we place greater trust in the reference data than in the model's prediction. 
% As the model gains knowledge during training, we use it to revise the ground-truth data progressively.  

We evaluate our self-correction method in both simulated and real-world noisy settings. 
We demonstrate that our method consistently outperforms baselines in both high- and low-resource datasets with different levels of misalignment noise. 
Moreover, we clearly show the gains are mainly from revising the misaligned samples while maintaining the performance of clean parallel data.
In the real-world noise setting, our self-correction method effectively handles naturally occurring noise in web-mined parallel datasets, e.g., ParaCrawl and CCAligned, achieving performance gains of up to 2.1 BLEU points across seven translation tasks and outperforming alternative methods, including pre-filters and truncation.

\section{Background}

\subsection{The Noisy World} 

Web-crawled parallel corpora are the primary training data source for machine translation models. 
However, parallel data crawled from public websites lack quality guarantees and contain different types of noise \citep{kreutzer-etal-2022-quality}, including wrong language, non-linguistic content, and semantic misalignment. 

The primary source of noise in parallel web-mined data is semantic misalignment \citep{khayrallah-koehn-2018-impact, kreutzer-etal-2022-quality, Ranathunga2024QualityDM}. 
For instance, \citet{khayrallah-koehn-2018-impact} analyzed the data quality of the raw ParaCrawl corpus, showing 77\% of the analyzed sentence pairs contain noise with half of them being misalignments. 
Wrong language and non-linguistic contents only account for a small portion and can be easily handled by filters, e.g., language identification toolkits \citep{herold-etal-2022-detecting}.
\citet{kreutzer-etal-2022-quality} extended the data quality analysis to pre-filtered web-mined datasets, e.g., WikiMatrix, CCAligned, noting that more than 50\% of data in both corpora are noisy while misalignment is the primary noise. 
% Even in the top-quality parallel data from the NLLB corpus, many misaligned sentences can still be found~\citep{Ranathunga2024QualityDM}. 

Overall, previous studies demonstrate the prevalence of noisy training data in web-mined corpora for machine translation and underscore the importance of noise-robust training, particularly in handling misaligned data.

\subsection{Learning in the Noisy World}

\subsubsection{Data Filter}
Data filtering is a straightforward way to mitigate the impact of the noise from translation corpora. 
Two kinds of filters are often used to ensure the semantic alignment in a sentence pair: 
(1) surface-level filters, e.g., removing sentence pairs that differ a lot in the source and target length;  
(2) semantic-level filters, relying on quality estimation models to score each sentence pair \citep{kepler-etal-2019-openkiwi, rei-etal-2020-COMET, peter-etal-2023-theres}. 
Other works consider misalignment detection as a ranking problem by training a classifier on annotated synthetic misaligned data \citep{briakou-carpuat-2020-detecting}.

In this paper, we mainly consider semantic-level filters for comparison, e.g., LASER \citep{Artetxe2018MassivelyMS} and COMET \citep{rei-etal-2020-COMET}, due to broad applicability and common usage.

% The simple data filter methods can remove the data noise before training, however, the filtered training examples might still be partially helpful for the model, especially in data-scarce scenarios. 

%(1) LASER \citep{Artetxe2018MassivelyMS} and , a sentence alignment pre-filter tool for web-mined corpora, i.e., CCAligned and WikiMatrix; 
%(2) COMET \citep{rei-etal-2020-COMET}, a widely-used quality estimation model for machine translation. 
%COMET, LASER, and XLM-R, as comparable baselines due to their strong ability to remove synthetic noise \citep{bane-etal-2022-comparison}. 
%Even with the simplicity of pre-filter methods, they are sensitive to the threshold o
%All these approaches rely on external models to select high-quality data before training. 
%However, our approach does not require such pre-selection but focuses on model training dynamics, which can be applied more broadly. 

\subsubsection{Training Robustness}
The primary limitation of data filters is that they discard entire training samples before training. 
To retain as much useful information as possible in noisy samples, several methods focus on mitigating their negative impact during model training.  
For instance, \citet{Wang2018DenoisingNM} proposed an online data selection approach that utilizes extrinsic trusted data to identify high-quality samples during training. 
Similarly, \citet{briakou-carpuat-2021-beyond} employed external semantic divergence tags to guide the training of the translation model. 
However, both of these approaches depend on external data or factors.

In this paper, we consider an alternative line of works, i.e., data truncation, which relies solely on the model's self-knowledge to ignore potential noise and further benefits the robustness of model training \citep{kang-hashimoto-2020-improved, Li2023ErrorNT}. 
For example, \citet{kang-hashimoto-2020-improved} use losses to estimate data quality, where tokens with high loss are considered as noise and will be ignored during training by setting their loss to zero. 
\citet{Li2023ErrorNT} further propose Error Norm Truncation, using the $l_2$ norm between the model’s prediction distribution and the one-hot ground-truth token distribution to measure data quality. 
Their method considers the model's prediction distribution of non-target tokens, providing a more accurate data quality measurement. 

% For example, \citet{kang-hashimoto-2020-improved, DBLP:conf/acl/GoyalXLD22, flores-cohan-2024-benefits} used loss to estimate the data quality by the model's predicted probability of the ground truth. 
% Ground-truth tokens with high loss are considered as noise and will be skipped during training by setting their loss to zero. 
% \citet{Li2023ErrorNT} further proposed Error Norm Truncation and used the $l_2$ norm between the model’s prediction distribution and the one-hot distribution of the ground-truth token to measure the data quality. 
% Their method also considers the model's probability distribution of non-target tokens, providing a more accurate measurement of data quality. 

However, there are two limitations of truncation methods: First, they \textit{ignore} the potential noisy training tokens from a specific training iteration, which overlooks the changes in the model's reliability during training.
Second, \textit{ignoring} can remove partially clean training information, which can be harmful for low-resource tasks.
In this paper, we go a step further and propose a self-correction method to gradually increase the trust of model prediction distributions to \textit{correct} rather than \textit{ignore} the ground-truth data during training.
Details are introduced in Section~\ref{sec:method}.

\section{An Empirical Study of Misalignment}
In this section, we investigate the primary source of noise, i.e., semantic misalignment, in a simulated setting. 
We first introduce a strategy to simulate realistic misalignment noise by controlling semantic similarity (Section~\ref{sec:simulated_noise}).
Next, we show the similarity of our simulated noise to real-world misalignment in terms of adequacy and its hard-to-detect nature (Section~\ref{real-world}).
Under our simulated noisy setting, we evaluate model-based metrics to distinguish data noise and highlight their potential limitations (Section~\ref{sec:fine-grained}).

\begin{table}[!t]
    \centering
    \begin{tabular}{p{0.5cm}p{6cm}}
        \hline 
         \texttt{\small{en}} & \small\textbf{Alcohol poisoning} is the biggest cause of death.  \\
         \texttt{\small{nl}} &  \small{\textbf{Jacht} is de belangriekste doodsoorzaak.} \newline \small\textit{\texttt{en:}\textbf{Hunting} is the biggest cause of death. }\\
         \hline
        \texttt{\small{en}} & \small{With Bravofly you can compare the flight prices Santa Cruz De La Palma of \textbf{over 400 of the most famous} airlines \textbf{in the world.}} \\
        \texttt{\small{de}} & \small{Bravofly findet für Sie sämtliche \textbf{Billigflüge Zürich} - Santa Cruz De La Palma der besten \textbf{europäischen} Billigfluggesellschaften.} \newline \small\textit{\texttt{en:} Bravofly finds all the \textbf{cheap} flights \textbf{Zurich} - Santa Cruz De La Palma from the best \textbf{European low-cost} airlines for you.}  \\
         \hline
         
    \end{tabular}
    \caption{\small Examples of misaligned sentences in the ParaCrawl dataset. \textbf{Bold} represents the misaligned meanings. \textit{Italic} text represents the English translation. }
    \label{tab:misalign}
\end{table}

\subsection{Simulating Misalignment Noise} \label{sec:simulated_noise}
To simulate misalignment, previous works \citep{bane-etal-2022-comparison,herold-etal-2022-detecting, Li2023ErrorNT} randomly shuffle target sentences of a clean parallel corpus. 
However, random shuffling noise can be easily removed by pre-filters based on length or semantic difference \citep{herold-etal-2022-detecting}, which oversimplifies the misalignments found in real-world web-mined corpora.
While \citet{briakou-carpuat-2020-detecting} proposed generating fine-grained misaligned targets by perturbing equivalent samples, e.g., deletion or replacement, their method does not guarantee the fluency and authenticity of the misaligned sentences.
%which are natural sentences mined by semantic alignment tool e.g. LASER (examples in Table \ref{tab:misalign}). 

%For instance, \citet{kreutzer-etal-2022-quality} show that a large portion of misaligned sentence pairs that share partial semantics still exists even after pre-filter tools, i.e., LASER, (examples in Table \ref{tab:misalign}) which show the challenge of detecting real-world misalignment.

To quantitatively analyze the impact of realistic misalignment noise, we designed a process to simulate real-world misalignment controlled by semantic similarity. 
The main idea is to select misaligned target sentences from a large pool of clean candidates that share partial semantics with the corresponding source sentences, where we use quality estimation models, e.g., LASER or COMET, to measure semantic similarity across languages. 

More specifically, given a source sentence and a large pool of target sentences, we first narrow down potential candidates based on the length differences and the word overlap ratio with the true parallel target to reduce computational costs. 
Then, the candidate with the highest semantic similarity score is selected as the final synthetic misaligned target. 
By this two-step process, we generate misalignment efficiently while maintaining shared semantics. 
Algorithm~\ref{alg:noise} provides a detailed description of our strategy. 
Examples of misaligned sentences generated using LASER (Misaligned-LASER) and COMET (Misaligned-COMET) can be found in Appendix~\ref{tab:misalign_examples}.

% Our algorithm aims to select a misaligned target from a large pool of potential misaligned target candidates within a clean corpus for a given source. 
% The selected misaligned target has to share certain semantics with the source.
% % while also containing misalignments. 

% To achieve this, we use quality estimation models, such as LASER or COMET, to measure semantic similarity between a source sentence and its potential misaligned candidates. 
% To reduce computational costs, we first narrow down potential misaligned candidates based on length differences and word overlap with the true parallel target. 
% Then, the candidate with the highest semantic similarity score is selected as the final synthetic misaligned target. 
% By applying this two-step process, we generate misalignment noise in an efficient yet realistic manner.
% Examples of misaligned sentences generated using LASER (Misaligned-LASER) and COMET (Misaligned-COMET) can be found in Appendix~\ref{tab:misalign_examples}.

\subsection{Real-World Misalignment}\label{real-world}

 \begin{table}[!t]
    \centering
     \resizebox{0.8\linewidth}{!}{
    \begin{tabular}{p{4.5cm}p{2cm}}
    \toprule   
    \textbf{Misaligned Types} & \textbf{Adequacy} \\ 
    \hline 
    {Real-World} & 3.1   \\
    {Misaligned-COMET} & 2.7\\
     {Misaligned-LASER } & 2.6 \\
     {Misaligned-Random } & 1.2 \\
    \bottomrule
    \end{tabular}}
    \caption{\small Adequacy (scale: 1–5) scores on simulated and real-world misaligned sentences. The real-world misaligned sentences are selected from ParaCrawl V7.0. Misaligned-COMET/LASER and real-world misaligned targets convey partial meanings with the sources. }
     \label{fig:adeq}
    
\end{table}

\begin{comment}
\begin{figure*}[!t]
    \includegraphics[width=\textwidth]{latex/image/loss.pdf}
\caption{\small \textit{loss} (above) and \textit{el2n} (below) distribution for clean and misaligned-\textit{LASER} noise samples during the training process (Epoch = 5, 10, 15, 30). \textcolor{red}{Red} distribution represents misaligned-\textit{LASER} noise and \textcolor{blue}{blue} distribution represents the clean data.}
\label{fig:el2n}
\end{figure*}
\end{comment}

\subsubsection{Adequacy}
To show the similarity of our simulated noise to real-world misalignment, we conduct a human evaluation of 200 simulated and real-world misaligned sentences, rating their \textit{Adequacy} (scale 1--5), which measures the meaning overlap between source and target. 
In Table~\ref{fig:adeq}, we show that both real-world misalignment and Misaligned-LASER/COMET (see Section~\ref{sec:simulated_noise}) have a relatively high adequacy score, above $2.5$, while random shuffled misaligned sentences only have an adequacy of $1.2$.  
This ensures our simulated misalignment contains only partial semantic overlaps as the real-world misalignment. 
%Moreover, our simulation process ensures the fluency of misaligned targets by selecting sentences from a clean corpus.  
Details of the human evaluation are in Appendix~\ref{sec:adequcy}.

\subsubsection{Hard-to-Detect Nature}\label{sec:filter_acc}
To show the hard-to-detect nature of our simulated noise, we investigate the noise detection ability of widely used pre-filters: COMET, LASER, Bi-Cleaner, and XLM-R. 
The details for each filter model are provided in Appendix~\ref{sec:filter}.

We calculate the noise detection accuracy of the data filters on a mixed set with the same amounts of clean and noisy data. 
For the clean data, we randomly sample 2,000 clean sentence pairs from the WMT2017 De$\rightarrow$En test set.
For Misaligned-Random, we randomly shuffle the order of target sentences in the sampled clean sentence pairs. 
For Misaligned-COMET and Misaligned-LASER, we use the same source sentences from the sampled clean data. 
We select the misaligned targets from another 200K target sentences in the training corpus based on Algorithm~\ref{alg:noise}. 
We score each sentence pair based on the filter models and determine a true ratio threshold based on the amounts of clean and noisy sentence pairs, here 1:1.
Sentence pairs with scores below this threshold are classified as noisy.

\begin{comment}
\begin{table}[!t]
    \centering
     \resizebox{0.9\linewidth}{!}{
    \begin{tabular}{p{3.2cm}ccc}
    \toprule  
          \textbf{Type} & \multicolumn{3}{c}{\textbf{Filter Accuracy}} \\
           \hline 
          & {COMET} & {LASER} &{XLM-R}   \\
    
            Misaligned-\textit{Random} &73\% & 76\%  & 70\%  \\
            Misaligned-\textit{LASER} & 52\% & 40\% & 55\%  \\
            Misaligned-\textit{COMET} & 50\% & 60\% & 52\%  \\
            
    \bottomrule
    \end{tabular}}
    \caption{\small Accuracy of data filters when distinguishing different misaligned noise from clean parallel data. }
    \label{tab:acc}
    \vspace{-0.3cm}
\end{table}
\end{comment}

\begin{figure}[!t]
    \centering
    \includegraphics[width=0.5\textwidth]{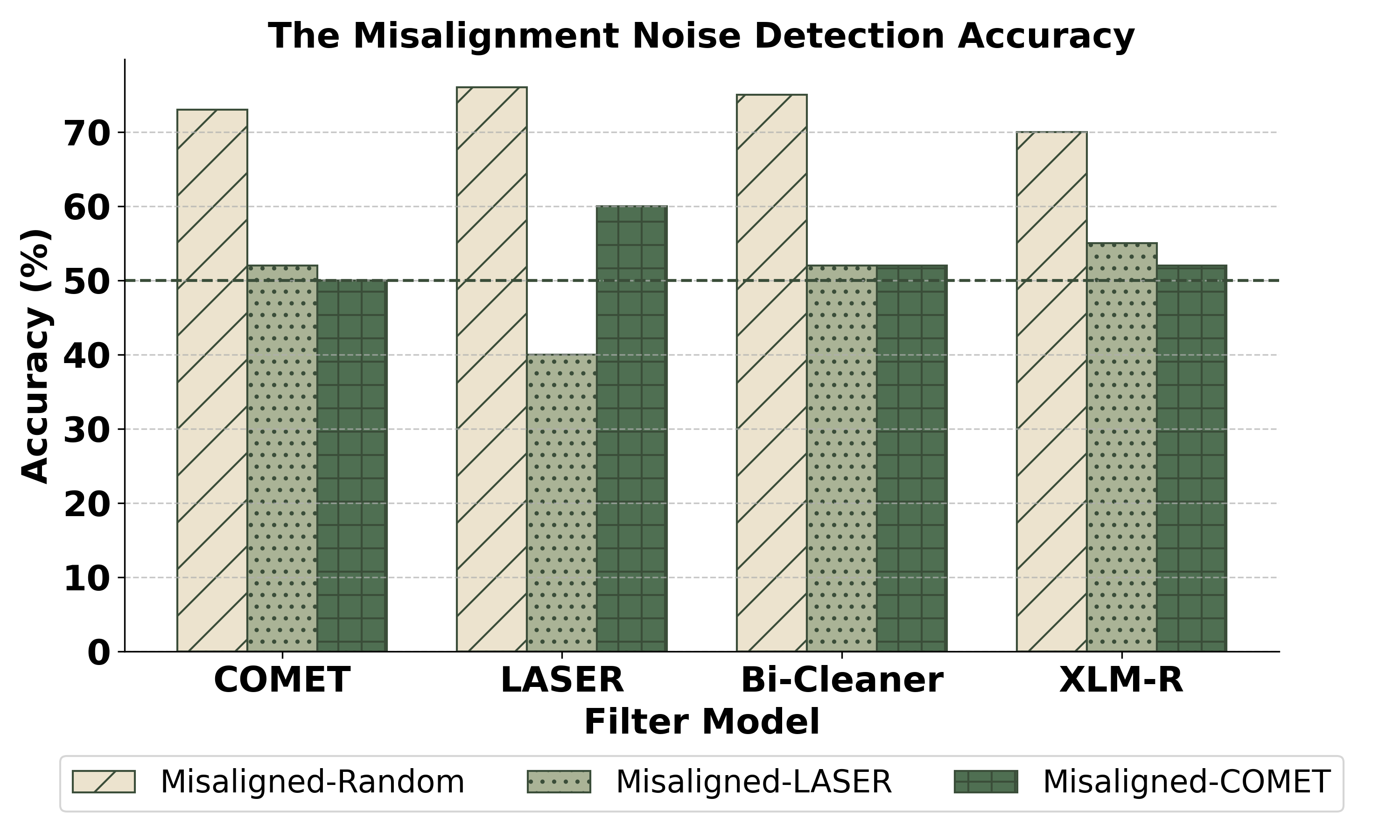}
   \caption{\small The accuracy of various data filters in distinguishing misaligned noise from clean parallel data. All four data filters \textbf{perform similarly to random guessing} (indicated by the black dashed line) on Misaligned-LASER/COMET.}
    \label{tab:acc}
\end{figure}

\begin{comment}
    \begin{table}[!t]
    \centering
     \resizebox{\linewidth}{!}{
    \begin{tabular}{p{3.2cm}cccc}
    \toprule  
          \textbf{Type} & \multicolumn{4}{c}{\textbf{Filter Accuracy}} \\
           \hline 
          & {COMET} & {LASER} &{Bi-Cleaner} & {XLM-R} \\
    
            Misaligned-\textit{Random} &73\% & 76\%  & 75\% & 70\% \\
            Misaligned-\textit{LASER} & 52\% & 40\% & 52\%  & 55\% \\
            Misaligned-\textit{COMET} & 50\% & 60\% &  52\% & 52\% \\
            
    \bottomrule
    \end{tabular}}
    \caption{\small Accuracy of data filters when distinguishing different misaligned noise from clean parallel data. All four data filters \textbf{perform nearly as random guessing} on Misaligned-LASER/COMET. }
    \label{tab:acc}
    
\end{table}
\end{comment}

Figure~\ref{tab:acc} shows the noise detection accuracy of data filters for different misaligned noise. 
First, all data filters have a relatively high detection accuracy for Misaligned-Random, particularly when using LASER, with an accuracy of 76\%. 
This questions previous assumptions \citep{khayrallah-koehn-2018-impact, Li2023ErrorNT} of the impact of misalignment noise on translation performance since most of them can be pre-filtered. 
However, our introduced noise, i.e., Misaligned-LASER and Misaligned-COMET, poses a challenge to all pre-filters, as the real-world misalignment does. 
%E.g., only around 50\% of filter accuracy for Misaligned-\textit{LASER} by COMET or XLM-R. 
%This shows misalignment noise with shared semantics requires a more fine-grained noise detector.  

Overall, we show the validity of our simulated noise in two aspects: (1) Adequacy, reflected in the similar level of shared semantics as the real-world misalignment; 
(2) Hard-to-Detect Nature, reflected in the low noise detection accuracy from widely used pre-filters. 
%Our simulated misalignment noise highlights the necessity of a fine-grained noise-handling way. 

\subsection{Fine-grained Misalignment Detection}\label{sec:fine-grained}

% Model-based metrics i.e., loss and error norm values, are used in data truncation methods \citep{kang-hashimoto-2020-improved, Li2023ErrorNT} to measure the data quality at token level. 

To measure the data quality during training, the token-level loss and error norm values are used in data truncation methods \citep{kang-hashimoto-2020-improved, Li2023ErrorNT}. 
Here, we evaluate their effectiveness under our simulated misalignment noise settings. 

% Loss measures the discrepancy between the ground-truth token and the model's point-wise prediction.
Loss measures the model's predicted probability of the ground-truth token. 
%but ignores the rich information in the probabilities of non-target tokens. 
%probability distribution of non-target tokens.
%Intuitively, training tokens with low probabilities are indicated as low-quality data. 
On the other hand, error norm value (\textit{el2n}) calculates the difference between the ground-truth (one-hot) distribution $\mathit{OH}(y_t)$ and the model's prediction distribution $p_{\theta}(\cdot|y_{<t}, x)$ (eq \ref{eq:el2n}).
%To include the non-target tokens' information, the error norm value (\textit{el2n}) considers the model's whole probability distribution, which calculates the difference between the ground-truth (one-hot) distribution $\mathit{OH}(y_t)$ and the model's prediction distribution $p_{\theta}(\cdot|y_{<t}, x)$ (eq \ref{eq:el2n}). 
Tokens with relatively high \textit{loss} or \textit{el2n} values are indicated as noise. 
%for each token $y_t$ during different stage of training:
\begin{equation}
    \mathit{el2n} = || p_{\theta}(\cdot|x, y_{<t}) - \mathit{OH}(y_t) ||_{2}.
    \label{eq:el2n}
\end{equation} 

\begin{figure}[!t]
    \centering
    % First row, first column
    \begin{subfigure}
    
        \includegraphics[width=0.45\linewidth]{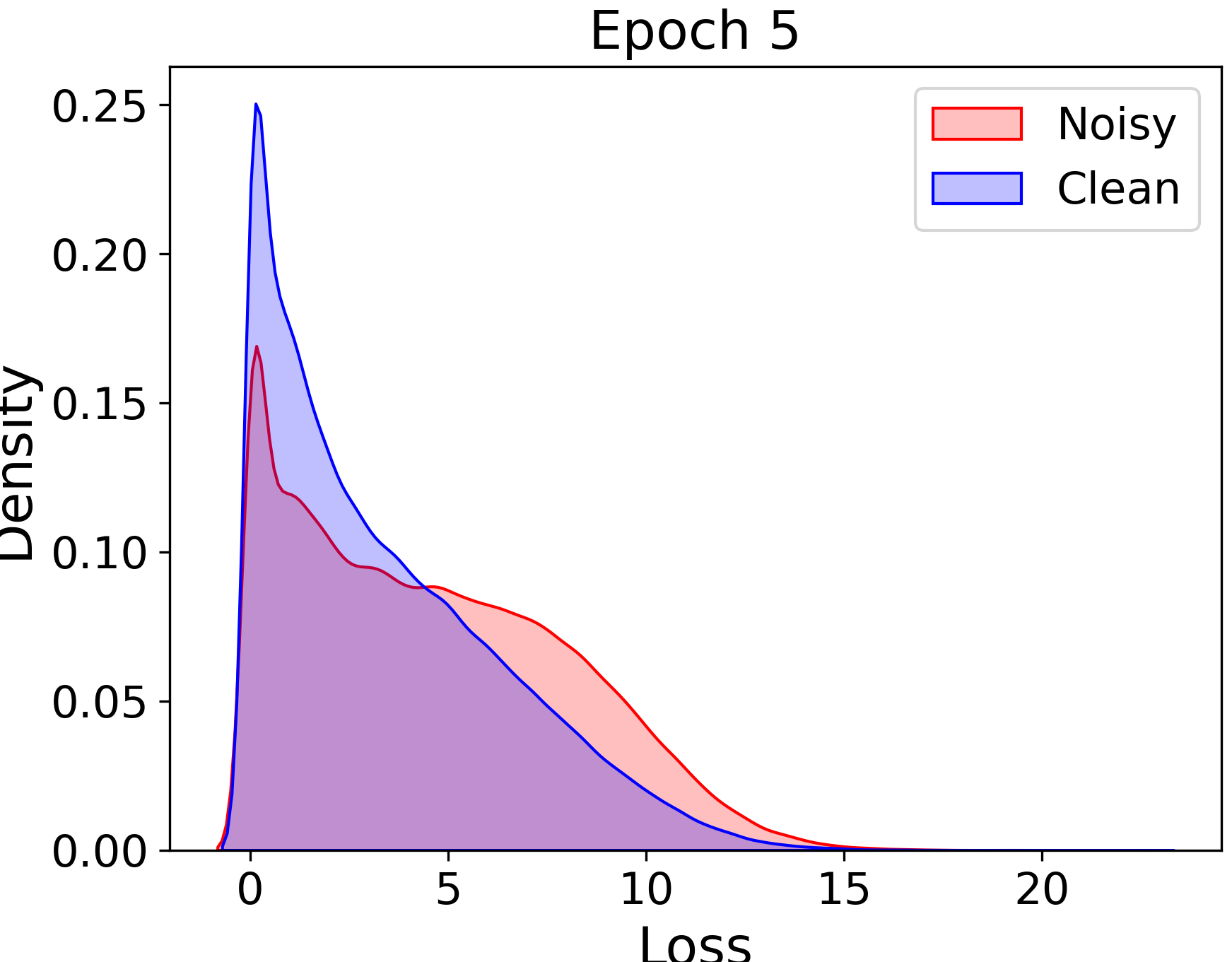}
        \label{fig:figure1}
    \end{subfigure}
    \hfill
    % First row, second column
    \begin{subfigure}
    
        \includegraphics[width=0.45\linewidth]{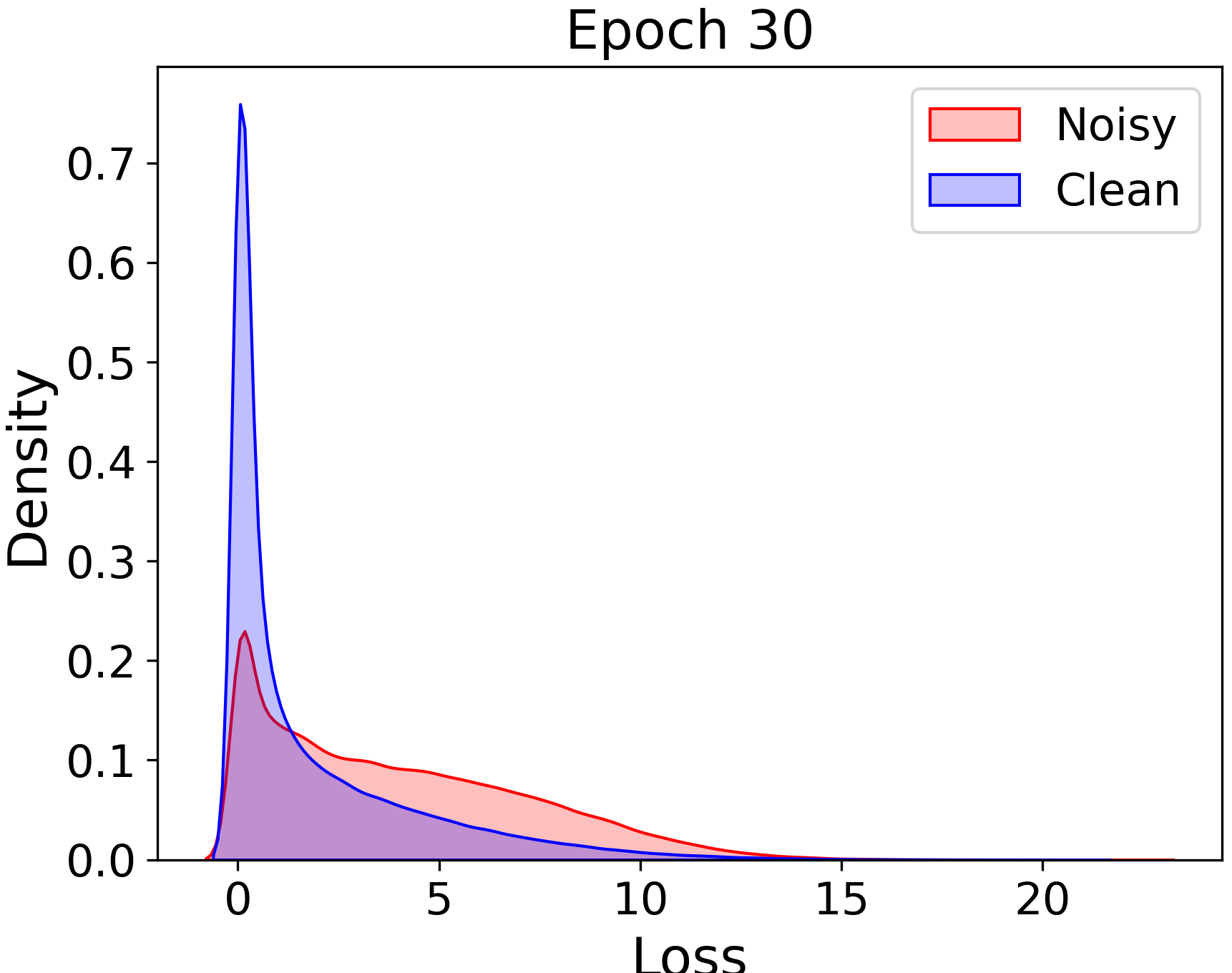}
        \label{fig:figure2}
    \end{subfigure}
    \vspace{0.3cm} % Space between rows

    % Second row, first column
    \begin{subfigure}
        
        \includegraphics[width=0.45\linewidth]{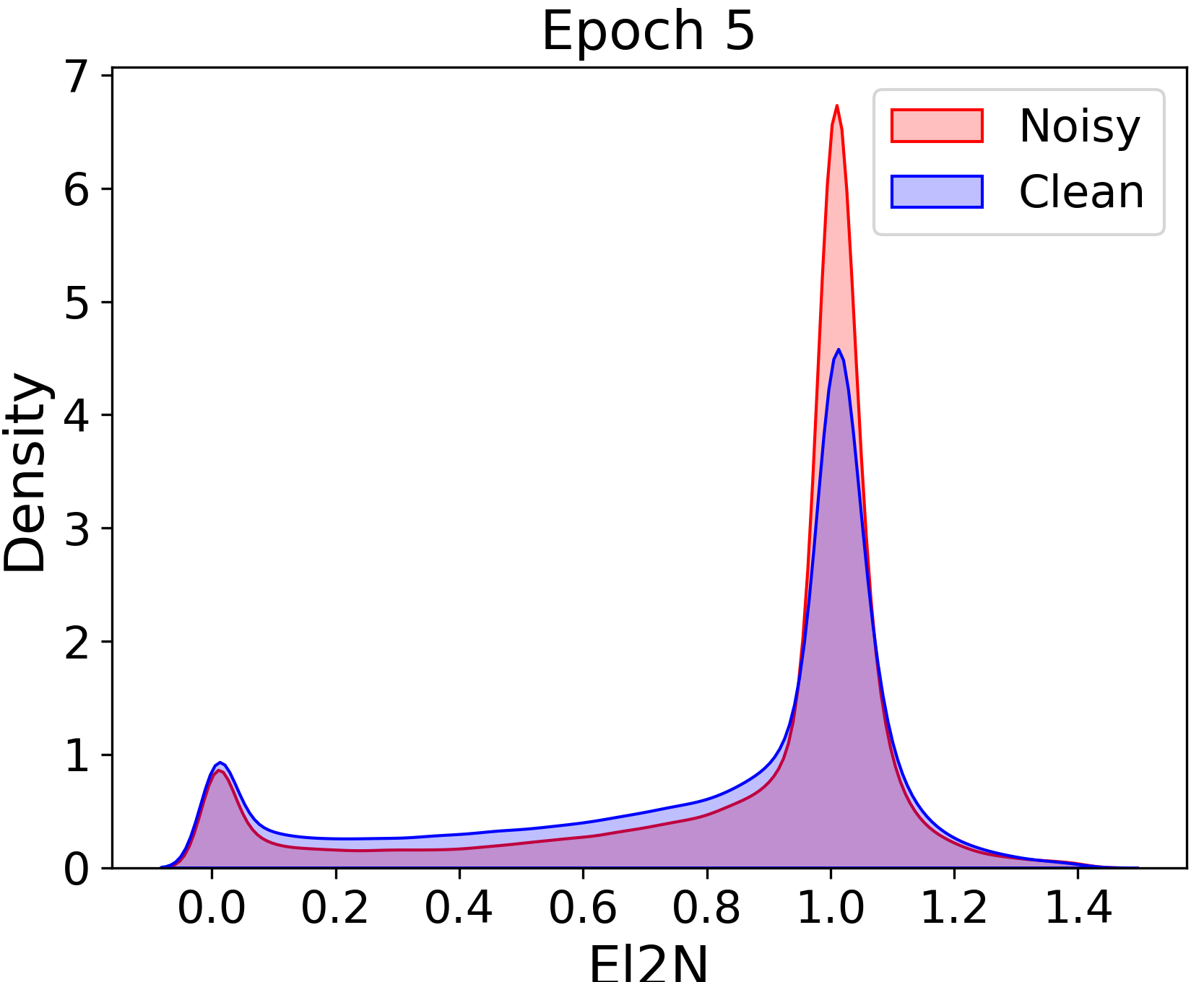}
        \label{fig:figure3}
    \end{subfigure}
    \hfill
    % Second row, second column
    \begin{subfigure}
       
        \includegraphics[width=0.45\linewidth]{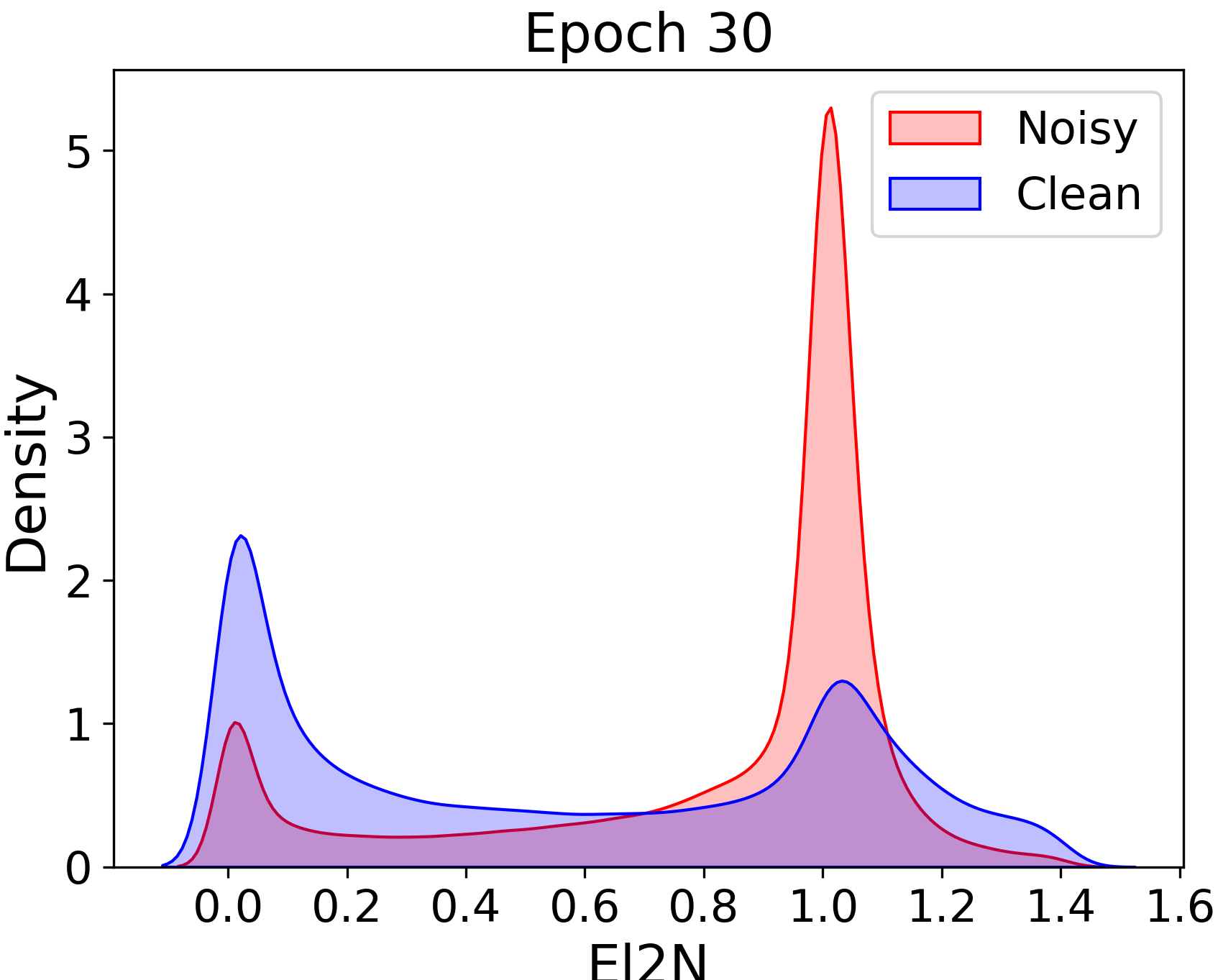}
        \label{fig:figure4}
    \end{subfigure}

    \caption{\small \textit{loss} (above) and \textit{el2n} (below) distribution for clean and Misaligned-LASER noise samples during the training process (Epoch = 5 and 30). \textcolor{red}{Red} distribution represents misaligned-\textit{LASER} noise and \textcolor{blue}{Blue} distribution represents the clean data. \textbf{As training progresses, \textit{el2n} distributions for clean and noisy data shift differently.} The distribution plots for the full training process are in the Appendix in Figure \ref{fig:el2n-more}. }
    \label{fig:el2n}
\end{figure}

We record the \textit{loss} and \textit{el2n} values for each token from 2,000 clean and Misaligned-{LASER} target sentences in the same data setting of Section~\ref{sec:filter_acc}. 
Figure \ref{fig:el2n} shows that clean and misaligned sentences have different \textit{loss} and \textit{el2n} distributions as  training time increases, from epoch $5$ to $30$. 
This shows the effectiveness of the model's self-knowledge for distinguishing hard-to-detect misalignment noise from clean sentences. 

Notably, the \textit{el2n} metric exhibits stronger distinguishability compared to \textit{loss}, underscoring the importance of considering the model's full prediction distribution. 
However, the noisy samples' \textit{el2n} distribution still partially shifts towards lower values during training, mainly due to the presence of clean tokens in the simulated misaligned sentences. 
To confirm that the shifted tokens in the noisy samples are truly clean, we provide a token-level annotation (see Appendix~\ref{sec:el2n-token}) to show that annotated misaligned tokens do have higher \textit{el2n} values (avg.\ $1.13$) than clean ones (avg.\ $0.32$).

Interestingly, we also observe that clean samples contain tokens with high \textit{el2n} values (see Table~\ref{tab:el2n-examples}). 
We hypothesize that these tokens might be difficult for the model to learn. 
Future work could further differentiate between hard-to-learn and noisy tokens and explore their respective impacts on the model's performance.

Overall, we point out two limitations of truncation methods relying on model-based metrics: 
First, they overlook the increasing reliability of model prediction by removing potential data noise from an early training time. 
Second, they cannot avoid ignoring clean but useful data. 
As mentioned, partial clean tokens still have high \textit{el2n} values. 

%Specifically, at the beginning of training, the \textit{el2n} distributions for clean and noise tokens are similar, with a large portion of tokens concentrating around $1.0$. 
%As the training time increases, the clean and noisy \textit{el2n} distributions progressively differ from each other: 
%which indicates the increasing reliability of the model's prediction distribution in distinguishing noisy data from clean ones. 
%The majority of clean tokens gradually have smaller \textit{el2n} values, while \textit{el2n} values from most of the noisy tokens are still large.
%This indicates the effectiveness of the model's prediction for distinguishing hard-to-detect data noise, showing the rationality of truncation methods to ignore training tokens by relying on metrics, e.g. \textit{el2n}. 

%This indicates the increasing reliability of the model's prediction distribution in distinguishing noisy data from clean ones. 
% In addition, the relatively large \textit{el2n} values for noisy data also indicate the model's `disagreement' with learning noisy ground-truth tokens. 
%This distribution difference motivates us to gradually leverage the model's prediction distribution to correct the ground-truth data. 

\section{Noise \textit{Self}-Correction}\label{sec:method}
To overcome the limitations of truncation methods in Section~\ref{sec:fine-grained}, we propose the self-correction method to gradually increase the trust of the model's prediction distributions to correct the supervision during training. 
Our method keeps the supervision signals from the training data to avoid clean training information loss and also progressively trusts a dynamic entropy state of the model's prediction to revise the data. 
Our work is in line with label correction in computer vision (discussed in Appendix~\ref{sec:label-correct}).

\paragraph{New Target.} Consider conditional probability models $p_{\theta}(y|x)$ for machine translation. 
Such models assign probabilities to a target sequence $y = \left(y_1, ..., y_T\right)$ by factorizing it to the sum of log probabilities of individual tokens $y_i$ from the vocabulary $V$.
At each training iteration, the model learns towards the ground-truth token distribution, one-hot $q(y_i)$, with a model prediction distribution $p_\theta(\cdot|x, y_{\textless{i}})$. 
In self-correction, we leverage the model prediction $p_\theta(\cdot|x, y_{\textless{i}})$ to revise the one-hot distribution $q(y_i)$ with the aim of learning towards a new target $\Bar{q}(y_i)$: 
\begin{equation}
    \Bar{q}(y_i) = (1-\lambda)q(y_i) + \lambda{{p_\theta(\cdot|x, y_{\textless{i}})}}
    \label{eq:qi}
\end{equation}
In this way, the new target $\Bar{q}(y_i)$ keeps the original supervision signal from the training data and the model's prediction. 
$\lambda$ denotes a weighting factor that determines how much to trust the model prediction. 

\paragraph{Dynamic Learning Schedule.}
We correlate $\lambda$ with a learning time function $\text{Time}(t)$ of training iteration $t$ and model entropy $H(p_\theta)$: 
\begin{equation}
     \lambda =  (1 - H(p_\theta)) \times \text{Time}(t) \\
\end{equation}
For $H(p_\theta)$, the model trusts its prediction more when it has a more confident prediction, i.e., lower entropy. 
For $\text{Time}(t)$, the model can trust its self-knowledge as training progresses.
We use a schedule \citep{NIPS2015_e995f98d} to increase $\text{Time}(t)$ as a function of the training iteration $t$ and $T$ as the number of total iterations. 
\begin{equation}
    \text{Time}(t) = \frac{1}{ 1 + \exp(\beta(\frac{t}{T} + \alpha))}
\end{equation}
where $\alpha$ and $\beta$ are hyper-parameters\footnote{We choose $\alpha$ and $\beta$ based on prior experiments, see Appendix~\ref{sec: hyper-para}. }.
%\footnote{ We set as $\alpha=-0.6$ and $\beta=-6$ to control the range of $\text{Time}(t)\in(0,1)$.}
%$T$ and $\beta$ are task-dependent and are tuned on validation sets.}.

In general, at the beginning of training, the model is not well-trained, and a small $\text{Time}(t)$ controls the model to rely more on the ground-truth data than its prediction. 
As training progresses, increasing $\text{Time}(t)$ allows the model to trust more in its reliable prediction. 

%We use a fixed $\tau$ to sharpen the model prediction in \citep{Wang2022ProSelfLCPS}
%Similar to the technique in \citet{Wang2022ProSelfLCPS}, we also leverage a softmax temperature $\tau$ to sharpen the model prediction distribution $ \Bar{p_\theta} = \frac{\exp(z_i/\tau)}{\sum_{j=1}^{N}\exp(z_j/\tau)}$ when learning towards the new target. 
%Instead of using a fixed value of $\tau$ to sharpen the model prediction in \citet{Wang2022ProSelfLCPS}, we propose a dynamic $\tau$ to learn towards a lowering entropy state of the model prediction according to the training time ratio $\frac{t}{T}$: 
%\begin{equation}
%    \tau = \frac{1}{ 1 + exp (\frac{t}{T})} + \mu
%\end{equation}
%where $\mu$\footnote{We set $\mu=0.1$ to control the range of $\tau$ in a low state between $\left(0.3, 0.6\right)$} determines the minimum value of $\tau$ to avoid over-confident model predictions. 
%In the dynamic setting, $\tau$ gradually decreases with the training: a higher $\tau$ at the early stage of training can prevent the model from converging and a smaller $\tau$ in the later stage makes the model more confident in its output. 

\begin{table*}[!t]
    \centering 
    \small
    \resizebox{\linewidth}{!}{
    \begin{tabular}{p{3cm}p{2cm}p{0.8cm}p{0.8cm}p{0.8cm}p{0.8cm}p{0.8cm}p{0.8cm}p{0.8cm}p{0.8cm}p{0.8cm}}
    \toprule
      & &  \multicolumn{3}{c}{\textbf{Misaligned-{LASER}}} &  \multicolumn{3}{c}{\textbf{Misaligned-{COMET}}} & \multicolumn{3}{c}{\textbf{Raw-Crawl Data}} \\
& & 10\% & 30\% & 50\% & 10\% & 30\% & 50\% & 10\% & 30\% & 50\%  \\
\cdashline{1-11}
%\multirow{1}{*}{\textbf{\small Baseline}} & \small \textit{with} noise  & 77.8\textsuperscript{*} & 77.0\textsuperscript{*} &  76.1\textsuperscript{*} &77.6\textsuperscript{*} & 76.5\textsuperscript{*} & 75.5\textsuperscript{*}  & 77.9\textsuperscript{*}  & 77.1\textsuperscript{*}  & 75.8\textsuperscript{*}   \\

 \multirow{1}{*}{\textbf{\small Baseline}} & \small \textit{with} noise  & 33.0\textsuperscript{*} & 31.7\textsuperscript{*} & 30.5\textsuperscript{*} & 33.1\textsuperscript{*} & 32.0\textsuperscript{*}  & 30.0\textsuperscript{*} & 33.0\textsuperscript{*}  & 31.5\textsuperscript{*}  & 29.6\textsuperscript{*}   \\
\hline 
\hline 
 \small \textbf{Oracle} & \small \textit{w/o} noise & 33.3 & 32.7 & 32.0  & 33.3 & 32.7 & 32.0 &33.3 & 32.7 & 32.0 \\
%\small \textbf{Oracle} & \small \textit{w/o} noise  & 79.5 & 79.0 & 78.6 & 79.5 & 79.0 & 78.6  &79.5   &  79.0 & 78.6   \\
\cdashline{1-11}
 %\small \textbf{Oracle} & \small \textit{w/o} noise & 33.3 & 32.7 & 32.0  & 33.3 & 32.7 & 32.0 &33.3 & 32.7 & 32.0 \\
\multirow{2}{*}{\textbf{\small Pre-Filter}} & \small LASER  & \underline{33.2}  & 31.4\textsuperscript{*}  & 30.0\textsuperscript{*} & 33.1\textsuperscript{*} & \textbf{32.6} & 30.2\textsuperscript{*} &  33.0\textsuperscript{*} & 31.6\textsuperscript{*}  & 30.0\textsuperscript{*}   \\
 & \small COMET  & 32.9\textsuperscript{*}  & 31.5\textsuperscript{*} & 30.4\textsuperscript{*} & 33.0\textsuperscript{*}  & 31.7\textsuperscript{*}  & 29.6\textsuperscript{*}  & 32.4\textsuperscript{*}  & 31.6\textsuperscript{*}  & 28.5\textsuperscript{*}  \\
 % & \small Bi-Cleaner  &  &   &   &   &   &   &   &  &    \\
 \cdashline{1-11}
\multirow{2}{*}{\textbf{\small Truncation}} & \small \textit{loss}  &  33.1\textsuperscript{*} &  31.4\textsuperscript{*} & 30.7\textsuperscript{*}    &  33.0\textsuperscript{*} &  31.2\textsuperscript{*} & 29.8\textsuperscript{*}   &  33.0\textsuperscript{*} &  \underline{31.8} & 29.9\textsuperscript{*} \\
 & \small \textit{el2n}  &  33.0\textsuperscript{*} & {31.9}\textsuperscript{*} & 31.0\textsuperscript{*} & 32.9\textsuperscript{*} & 31.8\textsuperscript{*} & 29.9\textsuperscript{*} & 33.0\textsuperscript{*} & 31.6\textsuperscript{*}  & 30.0\textsuperscript{*} \\
  % & \small el2n-\textit{threshold} & 33.3 & \underline{32.7} & 31.1\textsuperscript{*} & 33.3 & 32.3 &  29.8\textsuperscript{*} &33.0\textsuperscript{*}  & \underline{31.7} &30.2\textsuperscript{*}  \\
  \cdashline{1-11}
\multirow{2}{*}{\textbf{\small Self-Correction (Ours)}} & fixed $\tau=0.5$ & 33.1 & \textbf{32.9} & \underline{31.3} & \underline{33.2} & 32.4 & \underline{30.4}& \underline{33.4} & {31.7} & \underline{30.3} \\
& dynamic $\tau$  & \textbf{33.5} & \underline{32.3} & \textbf{31.4}  & \textbf{33.3} & \underline{32.5} & \textbf{30.6} &  \textbf{33.5} & \textbf{31.9}  & \textbf{30.4} \\
\bottomrule 
    \end{tabular}
    }
    \caption{\small SacreBLEU scores of high-resource De $\rightarrow$ En translation task with different types of noise. The BLEU score of the full clean training corpus (5.8M) De $\rightarrow$ En is 33.5.
    %The COMET score of the full clean training corpus (5.8M) De $\rightarrow$ En is 80.0.
    \textbf{Baseline} \textit{with} noise: represents the translation performance when injecting with 10\%, 30\%, 50\% of data noise. \textbf{Oracle} \textit{w/o} noise: represents the upper-bound translation performance when training with the remaining clean data, specifically 90\%, 70\%, 50\% of the data excluding the noise. \textbf{Bold} and \underline{Underline} represents the best and second best score. $*$ signifies that our self-correction method (dynamic $\tau$) is significantly better (p-value < $0.05$) than the comparing methods. The statistical significance results with paired bootstrap resampling are followed by \citep{koehn-2004-statistical}. COMET and Chrf++ scores are provided in Table \ref{tab:deen-COMET} in Appendix \ref{sec:chrf+}. }
    \label{tab:high-resource}
   
\end{table*}

\paragraph{Sharpen the Model Prediction.}
To overcome the overly uncertain model prediction on learning towards the new target in Equation~\ref{eq:qi}, we sharpen the model prediction distribution by controlling the softmax temperature 
$\tau$ in $\Bar{p_\theta} = \frac{\exp(z_i/\tau)}{\sum_{j=1}^{N}\exp(z_j/\tau)}$. 
We control $\tau$ in a dynamic way to vary it inversely with $\text{Time}(t)$. 
Therefore, $\tau$ gradually decreases as training time goes on: a higher $\tau$ at the early stage of training can prevent the model from converging and a smaller $\tau$ in the later stage makes the model more confident in its output. 

In Section~\ref{sec:exp}, we compare the performance of both fixed\footnote{We use fixed $\tau=0.5$ followed by \cite{Wang2022ProSelfLCPS}.} and dynamic $\tau$ to self-correct the data noise and also show the impact of different values of fixed $\tau$ on the performance in Appendix~\ref{sec:tau}.

\paragraph{Training.} 
After acquiring a new target $\bar{q}(y_i)$, derived from both the ground truth and the model's prediction, we obtain a new training objective based on maximum likelihood estimation (MLE). 
The following loss function is minimized for every training token over the training corpus $D$: 
\begin{equation}
    L_{\theta}(x,y) = \mathbb{E}_{y_i\sim{D}}\left[-\bar{q}(y_i)\log{{p}_\theta(\cdot|x, y_{\textless{i}}})\right]
\end{equation}

\section{Experiments}\label{sec:exp}

In this section, we investigate the effectiveness of our self-correction method for translation tasks in two experiment settings: simulated and real-world noisy settings. 
%(1) clean datasets with the presence of simulated noise (Section~\ref{sec:simulated}), and (2) real-world noisy corpus (Section~\ref{sec:real}). 
%We introduce our compared baseline systems in Section~\ref{sec:baseline}. 
For the simulated noisy setting (Section~\ref{sec:simulated}), we conduct experiments by injecting two types of noise, raw-crawl data and simulated misaligned noise, into a clean translation corpus. 
For the real-world noisy setting (Section~\ref{sec:real}), we perform experiments on two noisy web-mined datasets, i.e., ParaCrawl and CCAligned, across different language pairs.

\subsection{Comparing Systems}\label{sec:baseline}
We compare our self-correction method with the following comparing systems:\footnote{Note that all the models' details align with the corresponding baselines.}
%\paragraph{Oracle} \textit{w/o} noise, clean corpus training without injected noise which represents the upper-bound of translation performance. 
%\paragraph{Baseline.} \textit{with} noise, clean corpus training with an increasing level of different types of noise. 
%According to the real-world misaligned rate in web-mined datasets \citep{kreutzer-etal-2022-quality}, we injected $10\%-- 50\%$ of the misaligned noise or raw-crawl data of the clean data.  

\paragraph{Pre-Filtering.} 
We select two widely used data filters: LASER and COMET. 
We rank the training sentence pairs based on the scores calculated by the filter models. 
For the simulated noise experiments (Section \ref{sec:simulated}), we filter out the sentence pairs with the lowest scores before training, matching the size to the injected data noise. 
The training data size for pre-filter methods is 90\%, 70\%, and 50\% of the full training corpus when injecting with 10\%, 30\%, and 50\% of data noise.
For the real-world noise experiments (Section \ref{sec:real}), we filter out 20\% of the sentence pairs with the lowest scores.
%Details for the filter ratios are shown in Appendix \ref{sec:setup}. 
%We rank the training sentence pairs based on the scores calculated by the filter models. 
%For the simulated noise experiments (Section \ref{sec:simulated}), we filter out the sentence pairs with the lowest scores before training, matching the size to the injected data noise. 
%For the real-world noise experiments (Section \ref{sec:real}), we filter out 20\% of the sentence pairs with the lowest scores. 
%The training data size for pre-filter methods is 90\%, 70\%, and 50\% of the full training corpus with injecting 10\%, 30\%, and 50\% of data noise.

\paragraph{Truncation.} 
We compare two truncation methods: 
(1) \textit{loss} truncation \citep{kang-hashimoto-2020-improved}, 
(2) error norm value (\textit{el2n}) truncation \citep{Li2023ErrorNT}.
Following \citep{Li2023ErrorNT}, we choose the best result among three truncation fractions \{0.05, 0.1, 0.2\} for both \textit{loss} and \textit{el2n} truncation. 
The starting iteration to truncate data is set as 1500. 

%Details for the truncation fractions and thresholds are shown in Appendix \ref{sec:setup}. 
%Following \citep{Li2023ErrorNT}, we choose the best result among three truncation fractions \{0.05, 0.1, 0.2\} for both loss-\textit{fraction} and el2n-\textit{fraction} truncation. 
%We select the best result among three threshold values \{1.3, 1.35, 1.4\} for el2n-\textit{threshold} truncation.\footnote{Followed by \citep{Li2023ErrorNT}, the starting training iteration to truncate data is set as 1500. } 
%The starting training iteration to truncate data is set as 1500. 

\begin{table*}[!t]
    \centering 
    \small
    \resizebox{\linewidth}{!}{
    \begin{tabular}{p{3cm}p{2cm}p{0.8cm}p{0.8cm}p{0.8cm}p{0.8cm}p{0.8cm}p{0.8cm}p{0.8cm}p{0.8cm}p{0.8cm}}
    \toprule
      & &  \multicolumn{3}{c}{\textbf{Misaligned-{LASER}}} &  \multicolumn{3}{c}{\textbf{Misaligned-{COMET}}} & \multicolumn{3}{c}{\textbf{Raw-Crawl Data}} \\
& & 10\% & 30\% & 50\% & 10\% & 30\% & 50\% & 10\% & 30\% & 50\%  \\
\cdashline{1-11}
\multirow{1}{*}{\textbf{\small Baseline}} & \small \textit{with} noise  & 22.3 & 20.0\textsuperscript{*} & 18.0\textsuperscript{*} & 21.4\textsuperscript{*} & 18.7\textsuperscript{*}  & 14.2\textsuperscript{*}  & 22.3  & 21.0\textsuperscript{*}  & {19.0}\textsuperscript{*}  \\
\hline
\hline
\small \textbf{Oracle} & \small \textit{w/o} noise & 22.3 & 21.0  & 20.8  & 22.3 & 21.0 & 20.8 & 22.3 & 21.0 & 20.8 \\
\cdashline{1-11}
\multirow{2}{*}{\textbf{\small Pre-Filter}} & \small LASER  & 22.0\textsuperscript{*}  & 18.7\textsuperscript{*}  & 17.0\textsuperscript{*} &  21.1\textsuperscript{*} & 18.9\textsuperscript{*} & \textbf{16.3} &  21.0\textsuperscript{*} & 21.2\textsuperscript{*} & 19.2\textsuperscript{*}  \\
 & \small COMET  & 22.0\textsuperscript{*}  & 20.0\textsuperscript{*}  & 17.6\textsuperscript{*}  & 21.0\textsuperscript{*}  & 18.6\textsuperscript{*}  & 13.8\textsuperscript{*}  & 22.2  & 20.9\textsuperscript{*} & 18.9\textsuperscript{*}   \\
 %& \small Bi-Cleaner  &  &   &   &   &   &   &   &  &    \\
 \cdashline{1-11}
\multirow{2}{*}{\textbf{\small Truncation}} & \small \textit{loss}  &  22.1 &  20.5 & 17.9\textsuperscript{*}    & 20.0\textsuperscript{*} &  17.2\textsuperscript{*} & 14.2\textsuperscript{*}   &  22.2 & 21.1\textsuperscript{*} & 19.1\textsuperscript{*} \\
 & \small \textit{el2n}  &  22.0\textsuperscript{*} & 20.5 & 18.2\textsuperscript{*} & 21.1\textsuperscript{*} & 18.9\textsuperscript{*} & 14.3\textsuperscript{*} & 22.0\textsuperscript{*} &  21.3\textsuperscript{*}  & 19.2\textsuperscript{*} \\
  %& \small el2n-\textit{threshold} & 22.0\textsuperscript{*} & {20.6} & 19.4\textsuperscript{*} & \underline{22.0} &  {18.9}\textsuperscript{*}&  {14.5}\textsuperscript{*}& \underline{22.3} & \textbf{22.0} & {19.1}\textsuperscript{*}  \\
  \cdashline{1-11}
  
\multirow{2}{*}{\textbf{\small Self-Correction (Ours)}} & fixed $\tau=0.5$ & \textbf{22.4} & \textbf{21.2} & \underline{19.8} & \underline{21.7} & \underline{19.0} & 15.3 & \textbf{22.5} &\underline{21.5} & \textbf{19.9} \\
& dynamic $\tau$  & \underline{22.3} &  \underline{20.7} & \textbf{20.2}  & \textbf{22.1} &  \textbf{19.6} & \underline{16.2} & \underline{22.3} &  \textbf{21.9} &  \underline{19.6} \\
 \bottomrule
    \end{tabular}
    }
    \caption{\small SacreBLEU scores of low-resource En $\rightarrow$ Si translation task with different types of noise. The BLEU score of full clean training corpus (0.9M) En $\rightarrow$ Si is 22.5.
    %The COMET score of full clean training corpus (0.9M) En $\rightarrow$ Si is 82.0. 
    Chrf++ and COMET score are provided in Table \ref{tab:ensi-COMET} in Appendix \ref{sec:chrf+}.}
    \label{tab:low-resource}
  
\end{table*}

\subsection{Simulated Noisy World}  \label{sec:simulated}

\subsubsection{Experiment Setup}

We conduct experiments on both high- and low-resource translation tasks. 
We use the WMT2017 (German) De$\rightarrow$En news translation data as the high-resource task and En$\rightarrow$Si (Sinhala) from OPUS\footnote{\url{https://opus.nlpl.eu/}} as the low-resource task.

Following \citet{herold-etal-2022-detecting}, we inject noise by replacing a portion ({10\%, 30\%, 50\%}) of the clean training corpus with simulated misalignment noise or raw crawl data. 
The misalignment noise is generated by Algorithm~\ref{alg:noise} from the replaced portion of the clean corpus. 
The raw crawl data noise is randomly selected from the raw Paracrawl corpus\footnote{\url{https://paracrawl.eu/}}. 
Specifically, the raw crawl data provides a realistic test bed for noise-handling methods since it contains a mixture of naturally occurring noise, including misaligned sentences, wrong language, grammar errors, etc.

All translation models use the fairseq \citep{ott-etal-2019-fairseq} implementation of the Transformer-Big architecture for the high-resource task and Transformer-Base for the low-resource task. 
The full training details are shown in Appendix~\ref{sec:train-eval}.

\subsubsection{Results} \label{sec:result}

Tables~\ref{tab:high-resource} and \ref{tab:low-resource} show the high-resource De$\rightarrow$En and the low-resource En$\rightarrow$Si translation performance trained on the corpus with simulated misalignment or raw crawl data noise.
Overall, both noises negatively impact translation models, as shown by the performance drop with increasing noise levels.

First, we show that the pre-filter COMET fails to filter Misaligned-LASER noise, leading to a drop in translation performance in both high-resource and low-resource scenarios. This finding aligns with \citet{bane-etal-2022-comparison}, which demonstrates that COMET is weak at detecting misaligned segments. 
On the other hand, the pre-filter LASER is effective in handling Misaligned-COMET noise but only achieves modest gains when dealing with raw-crawl data noise.

Second, we demonstrate the effectiveness of leveraging the model's self-knowledge to detect data noise during training. 
Consistent with our findings in Section \ref{sec:fine-grained}, we show that using the \textit{el2n} metric yields better performance compared to using \textit{loss}. 
However, \textit{el2n} truncation still falls short in highly noisy environments ($50\%$). 
In such cases, the noisy datasets prevent the model from acquiring accurate knowledge, leading to incorrect data removal during training.

Our self-correction method overcomes the limitations of \textit{el2n} truncation by ``revising'' rather than ``ignoring'' data noise. 
This approach retains the ground truth supervision, preventing the loss of clean data information. 
This advantage is reflected in the superior performance of self-correction across low- and high-resource tasks in various noise settings. 
For instance, when injecting 50\% Misaligned-LASER noise into the En$\rightarrow$Si task, our self-correction method outperforms \textit{el2n} truncation by $2.0$ BLEU points.

Overall, our findings highlight the importance of utilizing the model's own predictions. This supports the hypothesis that training models solely on reference translations can limit performance, particularly when the reference is inferior to the model-generated translation \citep{xu2024contrastive}.

\begin{figure}[!t]
    \centering
    \includegraphics[width=0.45\textwidth]{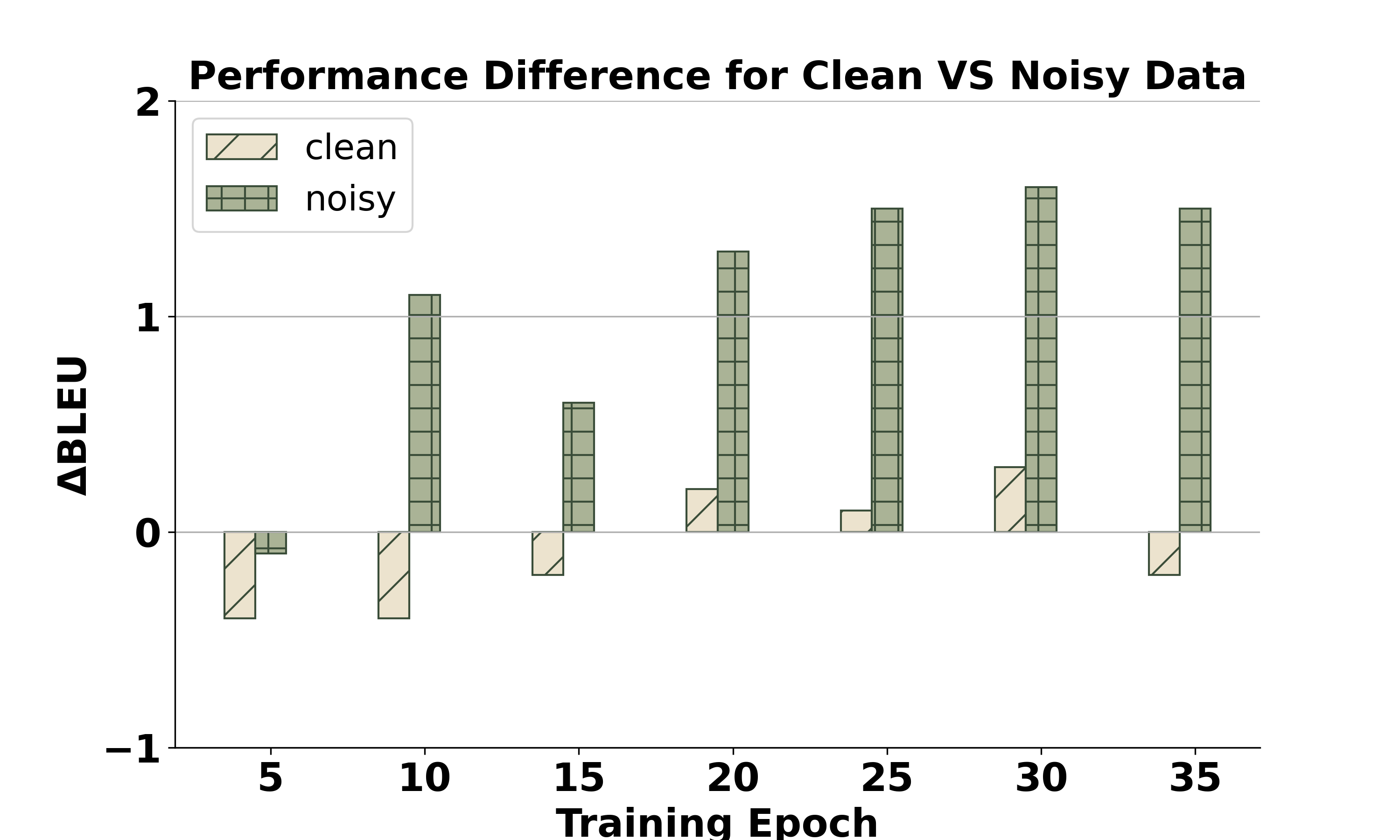}
    \caption{\small Performance differences between our self-correction method and baseline on noisy (Misaligned-LASER) and clean data for De$\rightarrow$En task with 30\% injected misaligned-LASER. \textbf{The effectiveness of our method mainly arises from improving the misaligned noisy data over clean ones.}}
    \label{fig:clean-noise}
\end{figure}

\begin{table*}[!t]
\centering
\resizebox{\linewidth}{!}{
\begin{tabular}{p{3.5cm}p{2.6cm}p{1.1cm}p{1.1cm}p{1.1cm}p{1.1cm}p{1.1cm}p{1.1cm}p{1.1cm}p{1.1cm}}
\toprule
  &    & en$\rightarrow$fr\textsuperscript{\heart} & en$\rightarrow$tr\textsuperscript{\Cross} & en$\rightarrow$es\textsuperscript{\Cross} & en$\rightarrow$be\textsuperscript{\Cross} & en$\rightarrow$si\textsuperscript{\heart} &  
  en$\rightarrow$sw\textsuperscript{\heart} & 
  en$\rightarrow$km\textsuperscript{\heart} & Avg. \\
   \textbf{Misaligned Rate (\%)} &  & 10\% & 44\%& 22\% & 10\% & 62\%& 11\% & 18\% & -  \\ 
   \textbf{Corpus Size (M)} &  & 5M &  5M & 5M & 1.1M& 210K& 130K & 60K  & - \\ 
  
\hline
\hline
 \textbf{Baseline} & & 41.1\textsuperscript{*} & 23.5\textsuperscript{*}  & 21.6\textsuperscript{*}  & 9.9\textsuperscript{*}  &   7.0\textsuperscript{*} & 13.0\textsuperscript{*} & 4.2\textsuperscript{*}  &  17.1  \\
 \cdashline{1-10} 
 \multirow{2}{*}{\textbf{Pre-Filter}} & {LASER} & 41.8\textsuperscript{*}  & 23.2\textsuperscript{*}&\underline{22.5} & 9.8\textsuperscript{*}&  6.6\textsuperscript{*} & 12.7\textsuperscript{*} & 3.8\textsuperscript{*} & 17.2 \\
 & {COMET} & 41.6\textsuperscript{*}  &23.7\textsuperscript{*} & 22.2\textsuperscript{*} & 9.6\textsuperscript{*} &  6.8\textsuperscript{*} & 12.5\textsuperscript{*} &4.0\textsuperscript{*} &  17.2\\
% & {Bi-Cleaner} &  &  & & & &  &  &  &   \\
\cdashline{1-10}
 \multirow{2}{*}{\textbf{Truncation}} & \textit{loss} & 41.2\textsuperscript{*} & 23.8\textsuperscript{*} & 21.9\textsuperscript{*} & 9.8\textsuperscript{*}  &   6.0\textsuperscript{*} & 12.5\textsuperscript{*}& 4.0\textsuperscript{*} & 17.0 \\
 & \textit{el2n} & 41.3\textsuperscript{*}  & \underline{23.9}\textsuperscript{*} & 22.0\textsuperscript{*} & 10.0\textsuperscript{*} & 6.0\textsuperscript{*} &  13.0\textsuperscript{*} & 4.5\textsuperscript{*}  & 17.2 \\
 %& {el2n-\textit{threshold}} & 41.2\textsuperscript{*}  & \underline{24.1} & 22.2\textsuperscript{*} & \underline{10.2}\textsuperscript{*} & 6.2\textsuperscript{*}  & 12.7\textsuperscript{*}& 4.2\textsuperscript{*}   & 17.3  \\
 \cdashline{1-10}
\multirow{2}{*}{\textbf{Self-Correction (Ours)}} & fixed $\tau=0.5$  &\underline{41.9} & 23.4 & 21.9 & \underline{10.1}   &  \underline{7.6}& \underline{14.7}& \underline{4.6} & \underline{17.9} \\
 & dynamic $\tau$  &  \textbf{42.3} &  \textbf{24.2} & \textbf{22.8}  &  \textbf{10.5}   & \textbf{7.8}  & \textbf{15.1} & \textbf{5.0} & \textbf{18.2} \\

\bottomrule

\end{tabular}
}   

\caption{\small SacreBLEU scores on real-world web-mined corpora. \textbf{Bold} and \underline{Underline} represents the best and second best score. {\Cross} denotes language pairs from CCAligned V1.0. {\heart} denotes language pairs from ParaCrawl V7.1. $*$ signifies that our self-correction method is significantly better (p-value < 0.05) than the baseline. The misaligned noise rate for different language pairs is reported from \citet{kreutzer-etal-2022-quality}. Chrf++ and COMET scores are provided in Table \ref{tab:real-COMET} in Appendix \ref{sec:chrf+}. } 
\label{tab:real-bleu}
\end{table*}

\subsubsection{The Sources of Improvements}
The previous section shows the benefits of the self-correction method in the presence of simulated misalignment noise.
To further investigate whether the improvements arise from addressing the misaligned data, we compare the differences in translation performance on clean and Misaligned-LASER data after applying the self-correction method. 

Specifically, we sample $1$K clean and Misaligned-LASER sentence pairs and report their BLEU score differences between the baseline and the self-correction model during training.
For Misaligned-LASER noisy data, BLEU scores are computed using the original parallel true references.
Figure \ref{fig:clean-noise} indicates that the effectiveness of our self-correction method primarily stems from improving the translation quality of misaligned data. 
Our method enhances performance on misaligned noisy data by up to 1.5 BLEU points during training, while its impact on clean data remains minimal.

\subsection{Real Noisy World} \label{sec:real}

\subsubsection{Experiment Setup}

We investigate two noisy web-crawled datasets: Paracrawl V7.1 and CCAligned V1.0. 
These two datasets exhibit varying semantic misalignment rates across different low- and high-resource language pairs \citep{kreutzer-etal-2022-quality}.
For each dataset, we select language pairs with varying levels of misaligned noise rate, from high- to low-resource. 
Training data details for the selected language pairs are shown in Appendix \ref{sec:langs}. 
%For Paracrawl, the language pairs are: en$\rightarrow$fr (French), en$\rightarrow$si (Sinhala), en$\rightarrow$sw (Swahili), and en$\rightarrow$km (Khmer). 
%For CCAligned, the language pairs are en$rightarrow$tr (Turkish), en$rightarrow$es (Spanish), and en$rightarrow$be (Belarusian). 
%For the high-resource language pairs: en$\rightarrow$fr, en$\rightarrow$tr, en$\rightarrow$es, we randomly sample 5M sentence pairs as the training corpus. 
%For medium and low-resource language pairs, we use the original corpus size. 
The validation and test sets for all tasks are from Flores101\footnote{\url{https://github.com/facebookresearch/flores}}.
We train for all tasks on the Transformer-Big \citep{10.5555/3295222.3295349} architecture. 
%We use the same comparing baseline systems as in Section \ref{sec:baseline}. 

\subsubsection{Results}
Table \ref{tab:real-bleu} shows the translation performance for two noisy web-crawled datasets, CCAligned V1.0 and Paracrawl V7.1, across language pairs with varying corpus size and misaligned rates.  

Similar to the findings in the simulated noise setting in Section \ref{sec:simulated}, we show that pre-filters and data truncation methods are limited to low-resource tasks with varying misalignment rates, e.g., en$\rightarrow$sw, en$\rightarrow$si, and en$\rightarrow$km, even degrading the translation performance. 
%though they can achieve performance gains for high-resource tasks. 
These two methods handle the data noise by removing or ignoring it; however, the noisy examples might still be partially helpful for the model, especially in data-scarce scenarios. 

In contrast, the self-correction method consistently outperforms other alternative methods, including pre-filters and truncation, with an overall improvement of $1.1$ BLEU, $1.7$ COMET, and $1.5$ ChrF++ points over the baseline. 
Specifically, self-correction shows superior performance in low-resource tasks, with up to $2.1$ BLEU and $2.4$ COMET points over the baseline for en$\rightarrow$sw task. 
This further emphasizes the effectiveness of using the model's self-knowledge to ``correct'' noise in real-world web-mined datasets.

\section{Conclusion}
In this paper, we aim to address the data quality issue in the web-mined translation corpora.
We show that the primary noise source in translation corpora, namely semantic misalignment, is hard to filter or handle by both widely used pre-filters and data truncation methods. 
To quantitatively analyze the impact of misalignment noise, we propose a process to simulate it controlled by semantic similarity, which reflects the partially shared meanings often found in misaligned sentence pairs from real-world web-crawled corpora. 

Under our simulated misalignment noise setting, we observe increasing reliability of the model's self-knowledge for detecting misalignments at the token level. 
Building on this, we propose \textit{self}-correction, which focuses on the model's training dynamics and revises the training supervision from the reference data by the model's prediction. Comprehensive experiments demonstrate the effectiveness of our approach on both simulated and real-world web-mined translation corpora.%, %achieving average performance gains of 1.7 COMET points across seven translation tasks. 
This performance outperforms alternative methods, including pre-filters and truncation methods. 
Moreover, we clearly show that the gains are mainly from revising the misaligned samples while maintaining the performance on clean data. 
Overall, our work provides a critical finding on the effectiveness of leveraging the model's predictions instead of solely relying on flawed reference data.

\section{Limitation}

First, we acknowledge the potential bias in our self-correction method, which could learn towards the noise due to its reliance on ground truth during the early training stages.
However, we believe this is not a significant issue because our method consistently demonstrates robust experimental results across different noise scenarios. 
Future work could explore modifications to mitigate this potential bias and enhance performance in diverse settings.

Second, our work aims at learning from a noisy training corpus, which might limit improvements when using high-quality training datasets.
Furthermore, the self-correction approach has shown promise in machine translation tasks, but another limitation is the unexplored potential for other natural language processing tasks, e.g., summarization or text generation. Future work should investigate the effectiveness of this approach across different downstream tasks.

\section*{Acknowledgments}

This research was funded in part by the Netherlands Organization for Scientific Research (NWO) under project number VI.C.192.080. We would like to thank Vlad Niculae, David Stap, Sergey Troshin and Evgeniia Tokarchuk for their useful suggestions. We would also like to thank the reviewers for their feedback.

%Third, we only focus on the primary source of noise, i.e., misaligned sentences, on machine translation. 
%Even we show the effectiveness of our approach on the real-world raw crawl data which contains naturally occurring data noise, future work can fine-grained evaluate our approach on other types of noise e.g., wrong language or misordered sentences. 

%\section*{Acknowledgments}

%This research was funded in part by the Netherlands Organization for Scientific Research (NWO) under project number VI.C.192.080. We would like to thank Vlad Niculae, David Stap, Sergey Troshin and Evgeniia Tokarchuk for their useful feedback and discussion.


\newpage

\bibliography{custom}

\newpage

\appendix

\section{Data Filters}\label{sec:filter}

For LASER \citep{Artetxe2018MassivelyMS}, it scores sentence pairs based on cross-lingual sentence embeddings. 
To calculate the LASER score for each sentence pair, we generate cross-lingual sentence embeddings using the pre-trained LASER model\footnote{\url{https://github.com/facebookresearch/LASER/blob/main/nllb/README.md}}.
The underlying system is trained as a multilingual translation system with a multi-layer bidirectional LSTM encoder and an LSTM decoder without information about the input language on the encoder. 
The output vectors of the encoder are compressed into a single embedding of fixed length using max-pooling, which is the cross-lingual sentence embedding resulting from the LASER model. 
The assumption is that two sentences with the same meaning but from different languages will be mapped onto the same embedding vectors. 
We calculate the LASER score followed by \citep{Chaudhary2019LowResourceCF}.  
The higher the LASER score, the more semantic similar between the source and target sentence.

For COMET, it is a neural framework for training machine translation evaluation models that can function as metrics \cite{rei-etal-2020-COMET}. 
Their framework uses cross-lingual pre-trained language modeling that exploits information from both the source input and the target reference to predict the target translation quality. 
We use the \texttt{reference-free wmt-20-qe-da} COMET model as the data filter to score each sentence pair in the training corpus. 

For Bi-Cleaner, it is a tool in Python that aims at detecting noisy sentence pairs in a parallel corpus. 
It indicates the likelihood of a pair of sentences being mutual translations.
Sentence pairs considered high-quality are scored near $1$, and those considered noisy are scored with $0$.
We use the multilingual model \texttt{bitextor/bicleaner-ai-full-en-xx} from HuggingFace\footnote{\url{https://huggingface.co/bitextor/bicleaner-ai-full-en-xx}} for the pre-filter for all language tasks. 

For XLM-R, it is a transformer-based multilingual masked language model pre-trained on text in 100 languages. 
We extract the sentence embeddings from the source and target with the model from \citet{Conneau2019UnsupervisedCR} and calculate their cosine similarity score as the XLM-R score. 

\section{Controlled Generated Misaligned Noise}

\subsection{Algorithm}

Algorithm \ref{alg:noise} shows the method to generate misaligned noise, controlled by two steps: (1) surface-level features control by word overlap and sentence length; (2) quality control by LASER or COMET. 

To save computational resources for calculating the LASER/COMET score for a source sentence with a chunk of target sentences, we first perform surface-level feature control (word overlap and length mismatch) to select a subset of misaligned target candidates. 
Word overlap is used as a filter to ensure that the misaligned targets share certain surface-level features with the true reference. 
Same for the length mismatch. 
%collectively determines how "similar" the misaligned target is to the true target.

To avoid overusing the selected misaligned target, we remove the selected target from the chunk of target sentences $T$. 
In our adequacy evaluation (shown in Appendix \ref{sec:adequcy} and Table \ref{tab:adequcy_q}), we also show that our misaligned sentences contain only partial meanings of the source sentences. 
This ensures a low likelihood that the selected misaligned target is a reasonable source sentence translation. 

\begin{algorithm*} 

\caption{Misaligned Noise Generation}\label{alg:noise}
\textbf{Input:} A chunk of parallel and de-duplicate clean data $D$ with N sentence pairs, source and target $(S, T)$; A threshold $k$ for selecting misaligned candidates; A quality controlled model $M \in \{\text{LASER}, \text{COMET}\}$ \\
\textbf{Output:} Misaligned data $\bar{D}$ with N sentence pairs source and misaligned target $(S, \Bar{T})$.

\begin{algorithmic}
    
\For{each source sentence $s_i$ in $S$}
    
    \State \colorbox{gray}{\textbf{Step 1: Surface-level Features Control}} 
    \State Initialize a list $L$ of misaligned candidates for $s_i$ 
\For{each target sentence $t_{j(j\neq i)}$ in $T$}
    
   \If{$ \operatorname{len}(L) < k$}
   
   \If {$|\operatorname{len}(t_j)-\operatorname{len}(s_i)|<3$ \textbf{and} $\text{word overlap ratio} (t_j, t_i) > 0.4$}
   \State Append $t_j$ to list $L$
    \EndIf
\EndIf
\EndFor
   
   \State \colorbox{gray}{\textbf{Step 2: Quality Control}} 
 \State Initialize a quality score list $Q$ 

 \For{each candidate $t_n$ in $L$}

 \State $\operatorname{score}(s_i, t_n) = M(s_i, t_n)$
 \State Append $\operatorname{score}$ to list $Q$
\EndFor
\State Select $t_k$ from $L$ with the highest score in $Q$ 
 \State Append the pair $(s_i, t_k)$ to the misaligned data $\bar{D}$ 
 \State Remove $t_k$ from targets $T$ to avoid $t_k$ over-reused 
\EndFor

\end{algorithmic}
\end{algorithm*}

\subsection{Misaligned Noise Samples}

Table \ref{tab:misalign_examples} shows the simulated misaligned samples of Misaligned-{LASER} and Misaligned-{COMET}. 
Overall, the simulated misaligned noise controlled by external models all share certain amounts of semantic meanings compared with the true reference. 

\begin{table}[!t]
    \centering
    \resizebox{0.9\linewidth}{!}{
    \begin{tabular}{p{3cm}p{7cm}}
    \toprule
         \textbf{SRC} & der Rat kam überein, dass die Kommission die Anwendung dieser Verordnung mit dem Ziel überwacht, etwaige Probleme möglichst schnell festzustellen und zu regeln. \\
         \cdashline{1-2}
         \textbf{REF} & the Council agreed that the Commission will keep under review the implementation of this Regulation with a view to detecting and addressing any difficulties as soon as possible.  \\
         \cdashline{1-2}
         \textbf{Mis-LASER} & the Commission has therefore acted wisely in exploring every possible avenue to guard against any difficulties and to prepare for any eventualities.\\
         \hline 
       \textbf{SRC} &  Brüssel , 17 März 2015 \\
       \cdashline{1-2}
        \textbf{REF} &  Brussels , 17 March 2015 \\
        \textbf{Mis-LASER} & Brussels , 4 May 2011 \\
        \hline 
     \textbf{SRC} & wann möchten Sie im Aeolos Hotel übernachten? \\
     \cdashline{1-2}
     \textbf{REF} & when would you like to stay at the Aeolos Hotel? \\
     \cdashline{1-2}
     \textbf{Mis-LASER} & when would you like to stay at the Leenane Hotel? \\
     \bottomrule
     \textbf{SRC} & buchen Sie Ihre Unterkunft in Edinburgh today! \\
     \cdashline{1-2}
     \textbf{REF} & book your accommodation in Edinburgh today! \\
      \cdashline{1-2}
      \textbf{Mis-COMET} & book your accommodation in Amsterdam today! \\
      \hline 
      \textbf{SRC} & wir akzeptieren folgende Kreditkarten:Visa, Maestro, Master Card, American Express, JBC, Dinners Club. \\
      \cdashline{1-2} 
      \textbf{REF} & We accept the following credit cards: Visa, Maestro, Master Card, American Express, JBC, Dinners Club. \\
      \textbf{Mis-COMET} & we accept payments by credit card (Visa, MasterCard, Diners Club), Paypal or transfer. \\ 
      \hline 
      \textbf{SRC} & Puchacz Puchacz Spa befindet sich in Niechorze , in einer schönen und malerischen Umgebung , ist lediglich 150m vom Meer entfernt und liegt in der Nähe des Liwia Łuża Sees . \\
      \textbf{REF} & Puchacz Puchacz Spa is located in Niechorze, in a beautiful and picturesque setting, only 150m from the sea and close to Lake Liwia Łuża. \\
      \textbf{Mis-COMET} & the Country Hotel Sa Talaia, surrounded by beautiful gardens is located close to San Antonio city and not far away from the historic city of Ibiza  \\
        \bottomrule
    \end{tabular}}
    \caption{Simulated Misaligned Sentences Samples}
    \label{tab:misalign_examples}
\end{table}

\subsection{Adequacy Evaluation} \label{sec:adequcy}

To evaluate the adequacy of the real world and our simulated misalignment noise, we design an annotation guide (see Table \ref{tab:adequcy_q}) to select the overlap meanings between a source sentence with the misaligned target. 
The simulated misaligned sentence pairs are constructed from the clean corpus WMT2017 De$\rightarrow$En, and the real-world misaligned sentences are selected from web-mined Paracrawl datasets.
The annotations were conducted by the two PhD students, who are also the authors of this paper, as volunteers without compensation.

\begin{table}[htbp]
    \centering
    \begin{tabular}{p{200pt}}
    \hline
           \small \textbf{Questionnaire}\\
           \hline
           \small Whether this target translation conveys the same meanings as the source sentence? \\
           $\circ$ \small all meanings  $\circ$ \small most meanings $\circ$ \small much meanings $\circ$ \small little meanings $\circ$ \small no meanings\\ 
           \hline
    \end{tabular}
    \caption{Questionnaire for human evaluation, where $\circ$ indicate single-item selection. From all meanings to no meanings, the adequacy score scales from 5--1.}
    \label{tab:adequcy_q}
\end{table}

\begin{figure*}[!t]
    \includegraphics[width=\textwidth]{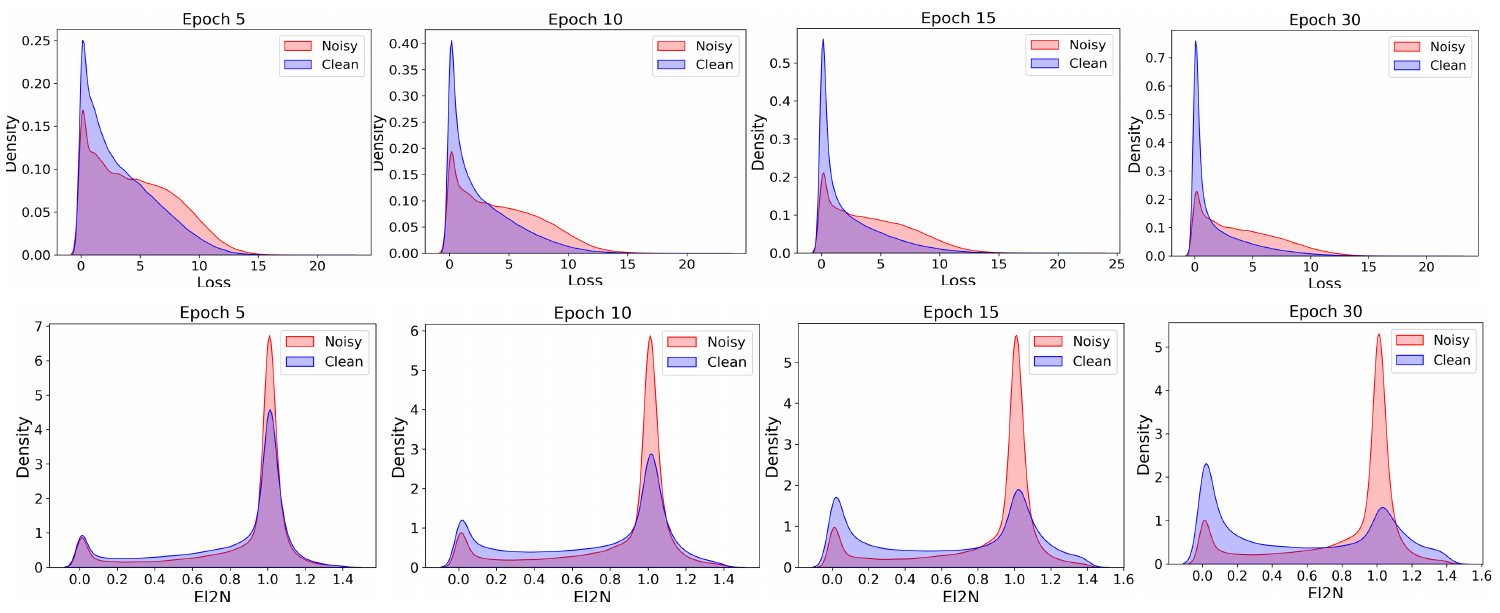}
\caption{\small \textit{loss} (above) and \textit{el2n} (below) distribution for clean and misaligned-\textit{LASER} noise samples during the training process (Epoch = 5, 10, 15, 30). \textcolor{red}{Red} distribution represents misaligned-\textit{LASER} noise and \textcolor{blue}{blue} distribution represents the clean data.}
\label{fig:el2n-more}
\end{figure*}

\subsection{Token-level Annotation} \label{sec:el2n-token}
We conducted a token-level annotation on 50 misaligned and clean sentences, resulting in 480 misaligned tokens and 1557 clean tokens. 
The annotators must label each token as ``clean'' or ``noisy'' given a source and a target sentence. 
The annotated misaligned and clean sentences are sampled from the sentences used in Section \ref{sec:fine-grained}. 
The annotations were conducted by the two PhD students, who are also the authors of this paper, as volunteers without compensation.

Table \ref{tab:token-el2n} shows that the average \textit{el2n} values for the misaligned tokens are higher than those for clean tokens, in both misaligned and clean samples, further confirming the effectiveness of leveraging the model's self-knowledge to distinguish data noise.
Moreover, we also find some clean tokens in clean target sentences do have higher \textit{el2n} values (shown in Table \ref{tab:el2n-examples}). 
We find that clean tokens with higher \textit{el2n} values tend to be difficult words for the model to learn, e.g., ``communication'' and ``developments''.

\begin{table}[htbp]
    \centering
     \resizebox{0.45\textwidth}{!}{
    \begin{tabular}{p{2cm}p{2cm}p{2cm}}
    \toprule
         \textbf{Misaligned} & \textbf{Clean-M } & \textbf{Clean-C}   \\
        \hline 
         1.13 & 0.32  & 0.37 \\ 
    \bottomrule
    \end{tabular}}
    \caption{Average \textit{el2n} values for annotated misaligned and clean tokens. Clean-M: Clean tokens from misaligned samples; Clean-C: Clean tokens from clean samples. }
    \label{tab:token-el2n}
\end{table}

\begin{table*}[htbp]
    \centering
     \resizebox{\textwidth}{!}{
    \begin{tabular}{cc}
    \toprule
        \textbf{SRC} & \_ganz \_entschieden \_möchte \_ich \_mich \_gegen \_den \_Ansatz \_der \_Kommission \_wenden \_, \_wie \_er \_in \_ihrer \_Mitteilung \_zum \_Ausdruck \_kommt \_.  \\
         \textbf{TGT} & \_I \_should \_also \_like \_to \_firmly \_contest \_the \_Commission \_\& apos ; s \_approach \_as \_presented \_in \_its \_communication \_.\\ 
         \textbf{High \textit{el2n}} & \_contest,  \_communication \\ 
         \hline 
            \textbf{SRC} & \_wir \_werden \_daher \_diesen \_Bericht \_unterstützen \_und \_das \_Thema \_auch \_weiterhin \_mit \_großer \_Aufmerksamkeit \_verfolgen \_.  \\
         \textbf{TGT} & \_we \_therefore \_support \_this \_report \_and \_will \_continue \_to \_closely \_monitor \_developments \_. \\ 
         \textbf{High \textit{el2n}} & \_closely, \_monitor, \_developments \\ 
         \hline 
           \textbf{SRC} & \_folglich \_muß \_bis \_zur \_Revision \_ein \_ausreichen der \_Zeitraum \_ver gehen \_.  \\
         \textbf{TGT} & \_we \_must \_therefore \_provide \_for \_a \_review \_after \_a \_sufficient \_period \_. \\ 
         \textbf{High \textit{el2n}} &\_therefore \\ 
    \bottomrule
    \end{tabular}}
    \caption{Clean sentence that contain tokens with high \textit{el2n} values. Here high \textit{el2n} represents the clean tokens have an \textit{el2n} value exceeding 1.35.} 
    \label{tab:el2n-examples}
\end{table*}

%\subsection{Other Misalignment Detection Methods}

%\citet{briakou-carpuat-2020-detecting} introduced a divergent mBERT model to detect semantic misalignment by training on synthetic annotated misaligned samples. 
%Here, we evaluate their model for detecting our simulated misaligned sentences for English into French data.
%Table x shows x, showing one possibility of fine-grained misalignment detection by training annotated divergent samples on external models. 

%However, we emphasize the major differences between their work and our self-correction method: (1) our work focuses on robustness training, which modifies the loss function and requires no external models or annotated data; (2) our method can be well applied to other tasks (e.g., summarization) and noise types, not only limited in semantic misalignment. 

\section{Self-Correction Method Design}

\subsection{Label Correction in Computer Vision} \label{sec:label-correct}

Our self-correction method is in line with the label correction method in Computer Vision \citep{Wang2022ProSelfLCPS, Lu2022SELCSL}. 
Both approaches are motivated by the idea of correcting data noise using a model's self-knowledge. 
However, we are the first work to apply this approach specifically in the text de-noise field. 

While other work \citep{kim2021selfknowledgedistillationprogressiverefinement} highlights another benefit of using the model's predictions to refine the target, i.e. regularization. 
However, we do not discuss this aspect in our paper. 
This is because we share different motivations. 
Our work primarily aims to improve the robustness of training to address the low-quality training data issues instead of regularizing the model.

\subsection{Hyper-Parameter Selection} \label{sec: hyper-para}
In $\text{Time}(t)$, $\alpha$ decides the inflection point, and $\beta$ adjusts the exponentiation's base and growth speed. 
Therefore, we fixed $\alpha = -0.6$ and conducted prior experiments to select $\beta$. 
Table \ref{tab:parameter-selection} provides the results of different $\beta$ under 30\% misaligned noise rations for high-resource and low-resource tasks. 
We select $\alpha = -0.6$ and $\beta = -6$ for our experiments. 
\begin{table}[htbp]
    \centering
    \resizebox{0.8\width}{!}{
    \begin{tabular}{c|c|c}
    \toprule 
        $\beta$ & High-resource & Low-resource \\
        \hline 
         -4 & 31.9  & 19.7 \\ 
         -5 & 32.0 & 20.0 \\
         \textbf{-6} & \textbf{32.3} & \textbf{20.3} \\
         -7 & 31.8 & 20.2 \\
         -8 & 31.4 & 20.0 \\
        \bottomrule
    \end{tabular}}
    \caption{Hyper-parameter Selection for $\beta$. We report the BLEU scores for different $\beta$ on high-resource task: De$\rightarrow$ En and low-resource task: En$\rightarrow$Si. }
    \label{tab:parameter-selection}
\end{table}

\subsection{The Impact of Sharpening Model Prediction.}\label{sec:tau}
\begin{figure}[!t]
    \centering
    \includegraphics[width=0.45\textwidth]{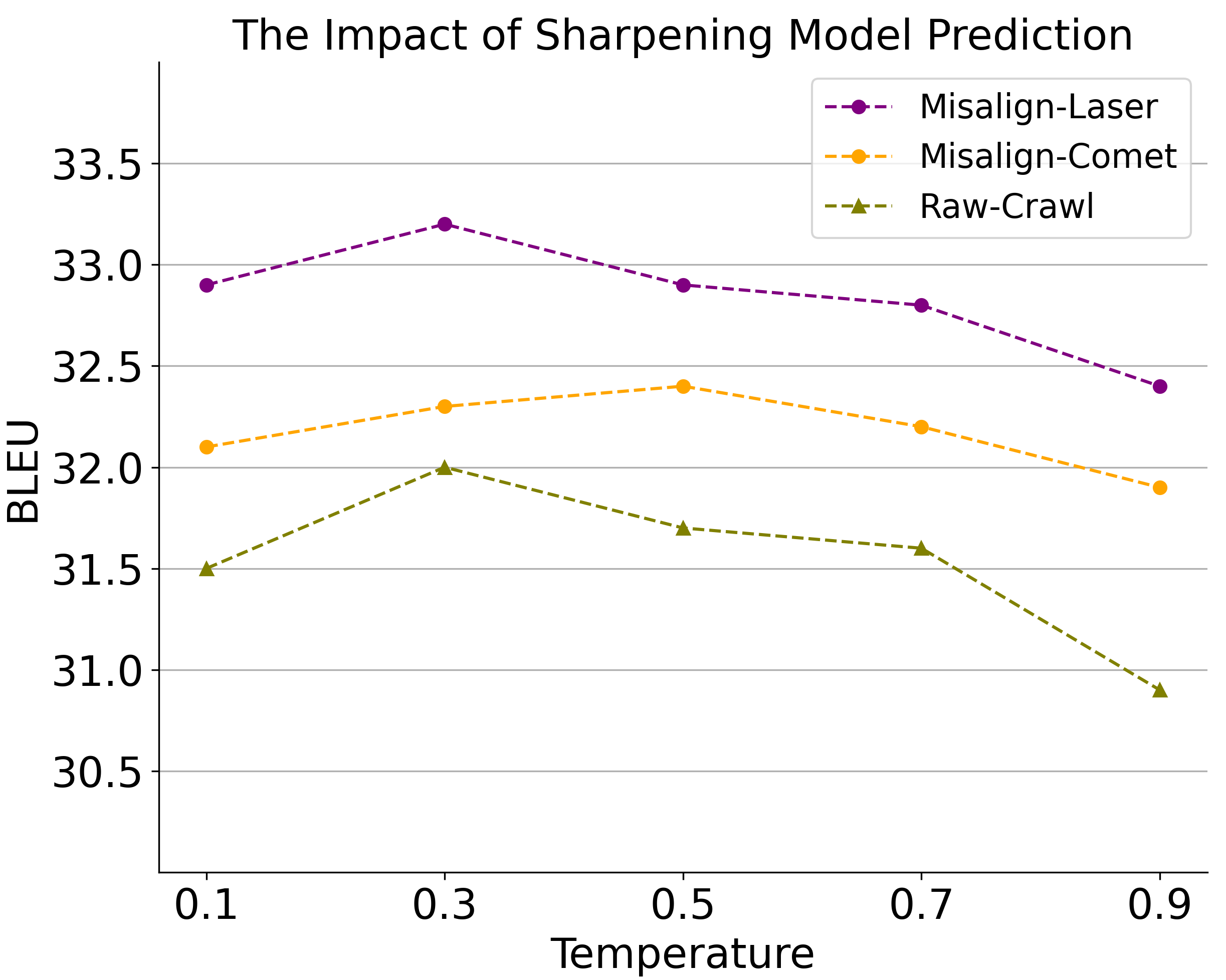}
    \caption{BLEU scores from the self-correction models on De$\rightarrow$En task with 30\% different types of injected noise with varying $\tau$.}
    \label{fig:emp-1}
\end{figure}

Here, we aim to analyze the impact of sharpening model prediction distribution, i.e., different fixed values of $\tau$, to correct the ground truth on translation performance. 
We train the self-correction models on De$\rightarrow$En task with 30\% of different types of noise, with varying values of softmax temperature $\tau$. 
From figure \ref{fig:emp-1}, we show that using sharpening model prediction distribution with a smaller $\tau$ achieves better translation performance for all noisy settings. 
However, the optimal $\tau$ varies when training with different types of noise and thus increases the difficulty of selecting a fixed $\tau$ for different scenarios. 
This motivates us to design a dynamic $\tau$, which varies automatically in a low range of entropy state over training time. 
The overall performance in both Section \ref{sec:simulated} and Section \ref{sec:real} by using a dynamic $\tau$ also shows its general applicability for different noise scenarios. 

\section{Training Details. } \label{sec:train}

%\subsection{Comparing Method Setup} \label{sec:setup}
%\paragraph{Data Filters. }
%We rank the training sentence pairs based on the scores calculated by the filter models. 
%For the simulated noise experiments (Section \ref{sec:simulated}), we filter out the sentence pairs with the lowest scores before training, matching the size to the injected data noise. 
%The training data size for pre-filter methods is 90\%, 70\%, and 50\% of the full training corpus when injecting with 10\%, 30\%, and 50\% of data noise.
%For the real-world noise experiments (Section \ref{sec:real}), we filter out 20\% of the sentence pairs with the lowest scores. 

%\paragraph{Truncation.}
%Following \citep{Li2023ErrorNT}, we choose the best result among three truncation fractions \{0.05, 0.1, 0.2\} for both loss-\textit{fraction} and el2n-\textit{fraction} truncation. 
%We select the best result among three threshold values \{1.3, 1.35, 1.4\} for el2n-\textit{threshold} truncation.
%Followed by \citep{Li2023ErrorNT}, the starting training iteration to truncate data is set as 1500. 

\subsection{Training and Evaluation}\label{sec:train-eval}
We follow the setup of the Transformer-base and Transformer-big models \citep{NIPS2015_e995f98d}. 
For each model, the number of layers in the encoder and in the decoder
is $N = 6$. 
We employ $h = 8$ parallel attention layers and heads for the Transformer-base. 
The dimensionality of input and output is $d_{\text{model}} = 512$, and the inner layer of feed-forward networks has dimensionality $d_{\text{ff}} = 2048$. 
We employ $h = 16$ parallel attention layers and heads for Transformer-big. 
The dimensionality of input and output is $d_{\text{model}} = 1024$, and the inner layer of feed-forward networks has dimensionality $d_{\text{ff}} = 4096$. 

All models are trained with the Adam optimizer \citep{Kingma2014AdamAM} for up to 500K steps for high-resource tasks and 100K steps for low-resource tasks, with a learning rate of 5e-4 and an inverse square root scheduler. 
A dropout rate of 0.3 and label smoothing of 0.2 are used.
Each model is trained on one NVIDIA A6000 GPU with a batch size of 25K tokens. 
We choose the best checkpoint according to the average
validation loss of all language pairs. 
The data is tokenized with the SentencePiece tool \citep{kudo-richardson-2018-sentencepiece}, and we build a shared
vocabulary of 32K tokens. 
For evaluation, we employ beam search decoding with a beam size of 5. 
BLEU scores are computed using detokenized case-sensitive SacreBLEU\footnote{nrefs:1|case:mixed|eff:no|tok:13a|smooth:exp|version:2.3.1}.

\subsection{Dataset Details}
\subsubsection{Simulated Noise Setting}
Table \ref{tab:simu-train} shows the training and evaluation dataset details for clean training corpus in simulated noisy experiments in Section \ref{sec:simulated}.

\begin{table}[htbp]
    \centering
    \resizebox{0.5\textwidth}{!}{
    \begin{tabular}{p{2.5cm}p{3cm}p{2cm}p{2cm}}
    \toprule
       Translation Task   & Training Source  & Dev Set  & Test Set  \\
       \midrule
       De$\rightarrow$En & WMT2017 (5.8M)  & NewsTest2016  & NewsTest2017 \\ 
       En$\rightarrow$Si & OPUS (0.9M)  & OPUS & OPUS  \\ 
    
        \bottomrule
    \end{tabular}}
    \caption{The clean training corpus and evaluation dataset details for experiments in Section \ref{sec:simulated}.}
    \label{tab:simu-train}
\end{table}

\subsubsection{Real-World Noise Setting} \label{sec:langs}

For Paracrawl, the language pairs are: en$\rightarrow$fr (French), en$\rightarrow$si (Sinhala), en$\rightarrow$sw (Swahili), and en$\rightarrow$km (Khmer). 
For CCAligned, the language pairs are en$\rightarrow$tr (Turkish), en$\rightarrow$es (Spanish), and en$\rightarrow$be (Belarusian). 
For the high-resource language pairs: en$\rightarrow$fr, en$\rightarrow$tr, en$\rightarrow$es, we randomly sample 5M sentence pairs as the training corpus. 
For medium and low-resource language pairs, we use the original corpus size. 

\section{Chrf++ and COMET Scores} \label{sec:chrf+}

Table \ref{tab:deen-COMET}, \ref{tab:ensi-COMET}, and \ref{tab:real-COMET} shows the COMET (\texttt
{Unbabel/wmt22-comet-da}) and Chrf++ scores for all experiments.

\begin{table*}[!t]
    \centering 
    \small
    \resizebox{\linewidth}{!}{
    \begin{tabular}{p{3cm}p{2cm}p{0.8cm}p{0.8cm}p{0.8cm}p{0.8cm}p{0.8cm}p{0.8cm}p{0.8cm}p{0.8cm}p{0.8cm}}
    \toprule
    & & \multicolumn{9}{c}{\textbf{COMET}} \\
     \cdashline{1-11}
      & &  \multicolumn{3}{c}{\textbf{Misaligned-{LASER}}} &  \multicolumn{3}{c}{\textbf{Misaligned-{COMET}}} & \multicolumn{3}{c}{\textbf{Raw-Crawl Data}} \\
& & 10\% & 30\% & 50\% & 10\% & 30\% & 50\% & 10\% & 30\% & 50\%  \\
\cdashline{1-11}
\multirow{1}{*}{\textbf{\small Baseline}} & \small \textit{with} noise  & 77.8\textsuperscript{*} & 77.0\textsuperscript{*} &  76.1\textsuperscript{*} &77.6\textsuperscript{*} & 76.5\textsuperscript{*} & 75.5\textsuperscript{*}  & 77.9\textsuperscript{*}  & 77.1\textsuperscript{*}  & 75.8\textsuperscript{*}   \\
\hline
\hline
\small \textbf{Oracle} & \small \textit{w/o} noise  & 79.5 & 79.0 & 78.6 & 79.5 & 79.0 & 78.6  &79.5   &  79.0 & 78.6   \\
\cdashline{1-11}
\multirow{2}{*}{\textbf{\small Pre-Filter}} & \small LASER  & 78.0\textsuperscript{*} & 76.9\textsuperscript{*} & 75.6\textsuperscript{*} & 78.2\textsuperscript{*}& \textbf{78.0} & 76.0\textsuperscript{*}  & 78.0\textsuperscript{*} & 77.8\textsuperscript{*}  & \underline{76.9}  \\
 & \small COMET  & 77.9\textsuperscript{*} & 77.5\textsuperscript{*} & 76.3\textsuperscript{*} & 77.5\textsuperscript{*} & 76.3\textsuperscript{*} & 74.0\textsuperscript{*}  & 78.0\textsuperscript{*}  & 76.8\textsuperscript{*}  & 75.6\textsuperscript{*}  \\
 \cdashline{1-11}
\multirow{2}{*}{\textbf{\small Truncation}} & \small \textit{loss} &78.3\textsuperscript{*}  & 76.5\textsuperscript{*} & 76.2\textsuperscript{*} & 78.0\textsuperscript{*}& 76.3\textsuperscript{*} & 75.0\textsuperscript{*}  & 78.0\textsuperscript{*}  & 77.2\textsuperscript{*}  & 76.6\textsuperscript{*}  \\
& \small \textit{el2n}  & 78.3\textsuperscript{*} & 78.3 & 76.5\textsuperscript{*} &78.1\textsuperscript{*} & 76.1\textsuperscript{*} & 76.0\textsuperscript{*}  & 78.2\textsuperscript{*}  & 77.5\textsuperscript{*}  & 76.2\textsuperscript{*}  \\
 % & \small el2n-\textit{threshold} & 78.5\textsuperscript{*} & \textbf{78.7} & 76.7 & 78.1\textsuperscript{*}&  76.6\textsuperscript{*}&  75.4\textsuperscript{*} &  78.5\textsuperscript{*} &  77.9\textsuperscript{*} &  76.6\textsuperscript{*} \\
  \cdashline{1-11}
  
\multirow{2}{*}{\textbf{\small Self-Correction (Ours)}} & fixed $\tau=0.5$ & \underline{79.0} & \underline{78.5} & \underline{76.8} & \underline{78.5} & {77.6} & \underline{76.2}  &  \underline{78.8} & \underline{78.1}  &  76.5 \\
& dynamic $\tau$  & \textbf{79.1}  & \textbf{78.6} & \textbf{77.0} & \textbf{78.7}  & \underline{77.7}& \textbf{76.6}  &   \textbf{79.0}&  \textbf{78.3}&  \textbf{77.0} \\
\bottomrule 
& & \multicolumn{9}{c}{\textbf{Chrf++}} \\
 \cdashline{1-11}
\multirow{1}{*}{\textbf{\small Baseline}} & \small \textit{with} noise  & 55.5\textsuperscript{*} & 54.9\textsuperscript{*} & 54.1\textsuperscript{*} & 55.1\textsuperscript{*} & 54.7\textsuperscript{*} & 52.5\textsuperscript{*} &  55.0\textsuperscript{*} &  54.9\textsuperscript{*} & 53.6\textsuperscript{*}  \\
\hline
\hline
\small \textbf{Oracle} & \small \textit{w/o} noise  & 57.2 & 56.9 & 55.5 &  57.2 &56.9  & 55.5   &   57.2 & 56.9  & 55.5   \\
\cdashline{1-11}
\multirow{2}{*}{\textbf{\small Pre-Filter}} & \small LASER  & 56.5\textsuperscript{*} & 54.5\textsuperscript{*} &53.4\textsuperscript{*}  &56.3 & \textbf{56.0} & 52.6\textsuperscript{*}  & 55.2\textsuperscript{*}  & 55.0\textsuperscript{*}  &  \underline{54.3}\textsuperscript{*} \\
 & \small COMET  & 56.0\textsuperscript{*} & 54.2\textsuperscript{*} & 53.0\textsuperscript{*} & 55.0\textsuperscript{*} & 54.2\textsuperscript{*} & 51.9\textsuperscript{*}  & 55.2\textsuperscript{*}  & 54.2\textsuperscript{*}  & 52.8\textsuperscript{*}  \\
 \cdashline{1-11}
\multirow{2}{*}{\textbf{\small Truncation}} & \small \textit{loss}  & 56.0\textsuperscript{*}&54.3\textsuperscript{*}  & 54.2\textsuperscript{*} & 55.5\textsuperscript{*}& 54.1\textsuperscript{*} & 52.0\textsuperscript{*}  & 55.5\textsuperscript{*}  & 55.0\textsuperscript{*}  & 54.0\textsuperscript{*}  \\
 & \small \textit{el2n}  &  56.1\textsuperscript{*}& 55.2\textsuperscript{*} & 54.2\textsuperscript{*} & 55.5\textsuperscript{*}& 55.0\textsuperscript{*} &  52.0\textsuperscript{*} & 56.2\textsuperscript{*}  &55.0\textsuperscript{*}   & 54.2\textsuperscript{*}  \\
  %& \small el2n-\textit{threshold} & 56.5\textsuperscript{*} &56.3  & \underline{54.6} & 56.2& 55.0\textsuperscript{*} & 52.8  & 56.2\textsuperscript{*}  & 55.5  & 54.0\textsuperscript{*}  \\
  \cdashline{1-11}
  
\multirow{2}{*}{\textbf{\small Self-Correction (Ours)}} & fixed $\tau=0.5$ &\underline{56.8} & \textbf{56.5}& \underline{54.3} & \textbf{56.6} & 55.2 & \underline{52.8}  & \underline{56.6} & \underline{55.5} & 54.0  \\
& dynamic $\tau$  & \textbf{56.9} & \underline{56.2} & \textbf{54.6} & \underline{56.4} & \underline{55.6} & \textbf{53.0}  &  \textbf{56.7} &\textbf{55.8} & \textbf{54.9}   \\
\bottomrule
    \end{tabular}
    }
    \caption{\small COMET and Chrf++ scores of high-resource De $\rightarrow$ En translation task with different types of noise. The COMET score of full clean training corpus (5.8M) De $\rightarrow$ En is 80.0. The Chrf++ score of full clean training corpus (5.8M) De $\rightarrow$ En is 57.2. $*$ signifies that our self-correction method is significantly better (p-value < 0.05) than the baseline. }
    \label{tab:deen-COMET}
  
\end{table*}

\begin{table*}[!t]
    \centering 
    \small
    \resizebox{\linewidth}{!}{
    \begin{tabular}{p{3cm}p{2cm}p{0.8cm}p{0.8cm}p{0.8cm}p{0.8cm}p{0.8cm}p{0.8cm}p{0.8cm}p{0.8cm}p{0.8cm}}
    \toprule
     & & \multicolumn{9}{c}{\textbf{COMET}} \\
     \cdashline{1-11}
      & &  \multicolumn{3}{c}{\textbf{Misaligned-{LASER}}} &  \multicolumn{3}{c}{\textbf{Misaligned-{COMET}}} & \multicolumn{3}{c}{\textbf{Raw-Crawl Data}} \\
& & 10\% & 30\% & 50\% & 10\% & 30\% & 50\% & 10\% & 30\% & 50\%  \\
\cdashline{1-11}
\multirow{1}{*}{\textbf{\small Baseline}} & \small \textit{with} noise  & 79.8\textsuperscript{*} & 79.0 & 77.8\textsuperscript{*} &79.7 & 75.9\textsuperscript{*} & 71.6\textsuperscript{*}  &  79.7\textsuperscript{*} &  79.5\textsuperscript{*} &  78.3\textsuperscript{*} \\
\hline
\hline
\small \textbf{Oracle} & \small \textit{w/o} noise  & 79.8 & 79.4  & 78.9 & 79.8 & 79.4 &78.9   & 79.8  & 79.4  &  78.9 \\
\cdashline{1-11}
\multirow{2}{*}{\textbf{\small Pre-Filter}} & \small LASER  & 79.5\textsuperscript{*} & 78.5\textsuperscript{*} & 77.0\textsuperscript{*} & 79.5\textsuperscript{*} &  76.2\textsuperscript{*} & \textbf{74.7}  & 79.8\textsuperscript{*}  & 79.8\textsuperscript{*}  &  79.0 \\
 & \small COMET  &79.6\textsuperscript{*}  & 78.8\textsuperscript{*} & 76.8\textsuperscript{*} &79.2\textsuperscript{*} & 76.0\textsuperscript{*} & 71.0\textsuperscript{*}  &  79.5\textsuperscript{*} & 79.0\textsuperscript{*}  & 77.8\textsuperscript{*}  \\
 \cdashline{1-11}
\multirow{2}{*}{\textbf{\small Truncation}} & \small \textit{loss}  & 79.9 & 78.4\textsuperscript{*} & 78.0 \textsuperscript{*} & 79.0\textsuperscript{*}& 75.6\textsuperscript{*} &71.2\textsuperscript{*}   &  {79.8}\textsuperscript{*} &  79.4\textsuperscript{*} & 78.6\textsuperscript{*}  \\
 & \small \textit{el2n}  & 80.1 & \underline{79.1} & 78.2\textsuperscript{*} &79.8 & 76.2\textsuperscript{*} &  72.3\textsuperscript{*} & 79.9  & 79.5\textsuperscript{*}  & 78.8\textsuperscript{*}  \\
 % & \small el2n-\textit{threshold} & 80.0 & 79.0 & \underline{78.5} &79.7 & 76.5\textsuperscript{*} & 73.5\textsuperscript{*}  &  80.0 & \underline{80.0}  & 79.0  \\
  \cdashline{1-11}
  
\multirow{2}{*}{\textbf{\small Self-Correction (Ours)}} & fixed $\tau=0.5$ & \textbf{80.3} & 79.0 & \underline{78.5} &\underline{79.9} & \underline{77.0} & 74.0  & \textbf{80.3}  & \underline{79.8}  & \textbf{79.5}  \\
& dynamic $\tau$  & \underline{80.1} &  \textbf{79.2}& \textbf{78.8} & \textbf{79.9}&  \textbf{77.1}&  \underline{74.6} &  \underline{80.2} &  \textbf{80.1} &  \underline{79.2} \\
 \bottomrule
 & & \multicolumn{9}{c}{\textbf{Chrf++}} \\
 \cdashline{1-11}
 \multirow{1}{*}{\textbf{\small Baseline}} & \small \textit{with} noise  &  35.7 & 34.0\textsuperscript{*} & 33.0\textsuperscript{*} & 34.9\textsuperscript{*} & 30.1\textsuperscript{*} & 24.2\textsuperscript{*} &   35.6\textsuperscript{*} & 34.0\textsuperscript{*}  & 32.7\textsuperscript{*}  \\
\hline
\hline
\small \textbf{Oracle} & \small \textit{w/o} noise  &35.9  & 34.6 & 34.2 & 35.9&34.6 & 34.2  & 35.9  & 34.6  & 34.2  \\
\cdashline{1-11}
\multirow{2}{*}{\textbf{\small Pre-Filter}} & \small LASER  & 35.4\textsuperscript{*} & 33.2\textsuperscript{*} & 32.5\textsuperscript{*} & 35.4\textsuperscript{*} & \underline{31.2} & \underline{28.0}   &  35.7 &  34.2\textsuperscript{*} & 33.0\textsuperscript{*}  \\
 & \small COMET  & 35.4\textsuperscript{*} & 33.5\textsuperscript{*} & 32.6\textsuperscript{*} & 33.6\textsuperscript{*} & 29.5\textsuperscript{*} &  23.8\textsuperscript{*} &  35.4\textsuperscript{*} &  33.8\textsuperscript{*} & 32.5\textsuperscript{*} \\
 \cdashline{1-11}
\multirow{2}{*}{\textbf{\small Truncation}} & \small \textit{loss}  & 35.8 & 33.6\textsuperscript{*} & 33.2\textsuperscript{*}  & 35.3\textsuperscript{*} & 30.2\textsuperscript{*} &  25.8\textsuperscript{*} &  35.7 &  34.2\textsuperscript{*} &  32.8\textsuperscript{*} \\
 & \small \textit{el2n}  & 35.6 & 34.1\textsuperscript{*} & 33.3\textsuperscript{*} & \underline{35.6} & 30.4\textsuperscript{*} & 26.0\textsuperscript{*}  &  35.6 &  34.1\textsuperscript{*} &   32.8\textsuperscript{*}\\
  %& \small el2n-\textit{threshold} & 35.7 & 34.2 & \underline{33.5}  & \textbf{36.0} & 30.8\textsuperscript{*} &  26.5\textsuperscript{*} &  35.6 &  \textbf{35.2} & 33.0\textsuperscript{*}  \\
  \cdashline{1-11}
  
\multirow{2}{*}{\textbf{\small Self-Correction (Ours)}} & fixed $\tau=0.5$ & \textbf{36.0} & \underline{34.3} & \underline{33.3}  & 35.5 & 31.0 &  27.0 & \textbf{36.0}  & \underline{34.8}  & \textbf{33.8} \\
& dynamic $\tau$  & \underline{35.8} & \textbf{34.4} & \textbf{33.6} & \textbf{35.8} & \textbf{31.5} & \textbf{28.3}   &  \underline{35.8} & \textbf{35.0}  & \underline{33.4}   \\
 \bottomrule
    \end{tabular}
    }
    \caption{\small COMET and Chrf++ scores of low-resource En $\rightarrow$ Si translation task with different types of noise. The COMET score of full clean training corpus (0.9M) En $\rightarrow$ Si is 82.0. The Chrf++ score of full clean training corpus (0.9M) En $\rightarrow$ Si is 37.0. $*$ signifies that our self-correction method is significantly better (p-value < 0.05) than the baseline. }
    \label{tab:ensi-COMET}
  
\end{table*}

\begin{table*}[!t]
\centering
\resizebox{0.95\linewidth}{!}{
\begin{tabular}{p{3.5cm}p{2.6cm}p{1.1cm}p{1.1cm}p{1.1cm}p{1.1cm}p{1.1cm}p{1.1cm}p{1.1cm}p{1cm}}
\toprule
   & & \multicolumn{8}{c}{\textbf{COMET}}\\ 
  \hline 
  &    & en$\rightarrow$fr\textsuperscript{\heart} & en$\rightarrow$tr\textsuperscript{\Cross} & en$\rightarrow$es\textsuperscript{\Cross} & en$\rightarrow$be\textsuperscript{\Cross} & en$\rightarrow$si\textsuperscript{\heart} &
  en$\rightarrow$sw\textsuperscript{\heart} &
  en$\rightarrow$km\textsuperscript{\heart} & Avg.  \\
   \textbf{Misaligned Rate (\%)} &  & 10\% & 44\%& 22\% & 10\% &  62\%& 11\% & 18\%  & - \\ 
   \textbf{Corpus Size (M)} &  & 5M &  {5M} & 5M & 1.1M&  210K& 130K & 60K & -  \\ 
  
\hline
\hline
 \textbf{Baseline} & & 80.0\textsuperscript{*}  &82.0\textsuperscript{*} & 76.5\textsuperscript{*}  & 68.3\textsuperscript{*} &    59.6\textsuperscript{*} & 59.0\textsuperscript{*} & 73.6\textsuperscript{*}  & 71.3 \\
 \cdashline{1-10} 
 \multirow{2}{*}{\textbf{Pre-Filter}} & {LASER}  & 81.0\textsuperscript{*} & 81.3\textsuperscript{*} & \underline{76.7}\textsuperscript{*} & 67.4\textsuperscript{*}& 59.7\textsuperscript{*} & 58.3\textsuperscript{*} & 73.6\textsuperscript{*} &  71.1  \\
 & {COMET} & 80.5\textsuperscript{*} &  81.0\textsuperscript{*} & 76.0\textsuperscript{*} &\underline{68.5}\textsuperscript{*} &  59.5\textsuperscript{*} & 58.1\textsuperscript{*} & 73.2\textsuperscript{*} & 71.0 \\
\cdashline{1-10}
 \multirow{2}{*}{\textbf{Truncation}} & \textit{loss} & 81.0\textsuperscript{*}  &82.2\textsuperscript{*} & 76.8\textsuperscript{*} & 67.6\textsuperscript{*}&  59.0\textsuperscript{*} & 58.8\textsuperscript{*} & 73.0\textsuperscript{*} &  71.2 \\
 & \textit{el2n} & 80.2\textsuperscript{*} & 82.1\textsuperscript{*} & 76.2\textsuperscript{*} & 68.6\textsuperscript{*} &  60.0\textsuperscript{*}& 58.6\textsuperscript{*} & 72.8\textsuperscript{*} & 71.2 \\
 %& {el2n-\textit{threshold}} &  \underline{81.4}\textsuperscript{*} & 82.3\textsuperscript{*} & \underline{76.8}\textsuperscript{*} & 68.7\textsuperscript{*}&   60.7\textsuperscript{*} & 58.9\textsuperscript{*} & 73.0\textsuperscript{*} & 71.7 \\
 \cdashline{1-10}
\multirow{2}{*}{\textbf{Self-Correction}} & fixed $\tau=0.5$   & \underline{81.2}  & \underline{82.5} & 76.4 &68.4 &  \underline{63.0}& \underline{61.0} & \underline{74.5} &  \underline{72.4} \\
 & dynamic $\tau$  & \textbf{81.6} &  \textbf{83.0} & \textbf{77.9}     & \textbf{68.9}  &  \textbf{63.6} & \textbf{61.4} &  \textbf{75.0} & \textbf{73.0} \\
 
\hline
 & & \multicolumn{8}{c}{\textbf{ChrF++ }}\\ 
 \hline 
 \textbf{Baseline} & & 67.3\textsuperscript{*} &  54.8\textsuperscript{*}  & 49.1\textsuperscript{*}  & 36.5\textsuperscript{*}  &   20.2\textsuperscript{*} & 37.9\textsuperscript{*} & 15.6\textsuperscript{*} & 40.2 \\
 \cdashline{1-10} 
 \multirow{2}{*}{\textbf{Pre-Filter}} & {LASER} & 67.9\textsuperscript{*} &  54.6\textsuperscript{*} & \underline{49.6}\textsuperscript{*} & 36.1\textsuperscript{*} & 21.7\textsuperscript{*} & 37.5\textsuperscript{*} & 14.7\textsuperscript{*} &  40.3 \\
 & {COMET} & 67.6\textsuperscript{*} &  54.3\textsuperscript{*}&49.6\textsuperscript{*}  &36.2\textsuperscript{*} & 20.6\textsuperscript{*}  & 37.2\textsuperscript{*} &15.0\textsuperscript{*} & 40.1 \\
\cdashline{1-9}
 \multirow{2}{*}{\textbf{Truncation}} & \textit{loss} & 67.4\textsuperscript{*}  & 55.2  & 49.2\textsuperscript{*}  & 36.3\textsuperscript{*} & 20.0\textsuperscript{*} & 37.9\textsuperscript{*} & 13.4\textsuperscript{*}  & 39.9 \\
 & \textit{el2n} & 67.6\textsuperscript{*}  &  \underline{55.2} & 49.5\textsuperscript{*}  & 36.5\textsuperscript{*} &  20.6\textsuperscript{*} & 37.3\textsuperscript{*} & 13.0\textsuperscript{*} & 40.1 \\
 %& {el2n-\textit{}} & 67.3\textsuperscript{*}  & \underline{55.3} & 49.3\textsuperscript{*} & 36.8\textsuperscript{*} & 20.4\textsuperscript{*} & 37.9\textsuperscript{*} & 13.9\textsuperscript{*} &  40.1\\
 \cdashline{1-9}
\multirow{2}{*}{\textbf{Self-Correction}} & fixed $\tau=0.5$  & \underline{68.0}  & 54.9 & 49.6 & \underline{36.8}  &  \textbf{24.0} & \underline{41.0} & \underline{16.5} & \underline{41.5} \\
 & dynamic $\tau$  &  \textbf{68.2}  & \textbf{55.4} & \textbf{50.0}  & \textbf{37.2}  & \underline{22.2}& \textbf{42.3} & \textbf{16.8 } &  \textbf{41.7}\\

\bottomrule

\end{tabular}

}   
\caption{\small COMET and Chrf++ scores on real-world web-mined corpora. For pre-filter methods, we remove 20\% of the training samples with the lowest scores. {\Cross} denotes language pairs from CCAligned V1.0. \heart denotes language pairs from ParaCrawl V7.1. The misaligned noise rate for different language pairs is reported from \citet{kreutzer-etal-2022-quality}. $*$ signifies that our self-correction method is significantly better (p-value < 0.05) than the baseline. }\label{tab:real-COMET}
\end{table*}

\end{document}